\documentclass[runningheads]{llncs}

\usepackage{eccv}

\usepackage{eccvabbrv}

\usepackage{graphicx}
\usepackage{booktabs}
\usepackage{ulem}
\usepackage[accsupp]{axessibility} 
\usepackage{multirow}
\usepackage{longtable}
\usepackage{amsmath}
\usepackage{amssymb}
\usepackage{pifont}
\newcommand{\customfootnotetext}[1]{%
  \begingroup
    \renewcommand{\thefootnote}{}
    \footnotetext{#1}%
  \endgroup
}

\usepackage[breaklinks,colorlinks,citecolor=eccvblue]{hyperref}

\usepackage{orcidlink}
\usepackage{marvosym}
\begin{document}

\title{Edit in 2D, Verify in 3D: Reinforcement Learning for Multi-view Consistent Scene Editing}

\titlerunning{Edit in 2D, Verify in 3D}

\author{Jiyuan Wang\inst{1,2,3} \and
  Chunyu Lin\textsuperscript{1,\Letter} \and
  Lei Sun\textsuperscript{2,\ding{81}}\and
  Zhi Cao\inst{1} \and Yuyang Yin\inst{1} \and Lang Nie\inst{4} \and Zhenlong Yuan\inst{2} \and Xiangxiang Chu\inst{2} \and Yunchao Wei\inst{1} \and Kang Liao\inst{3} \and Guosheng Lin\textsuperscript{3,\Letter}
  }
\authorrunning{Jiyuan Wang et al.}
\institute{\textsuperscript{1}BJTU, \textsuperscript{2}AMap Alibaba Group, \textsuperscript{3}NTU, \textsuperscript{4}CQUPT\\
\email{\{wangjiyuan,cylin\}@bjtu.edu.cn}}

\maketitle
\enlargethispage{2\baselineskip}
 \begin{figure}[!ht]
    \centering
    \vspace{-1.5em}
   \includegraphics[width=1\linewidth]{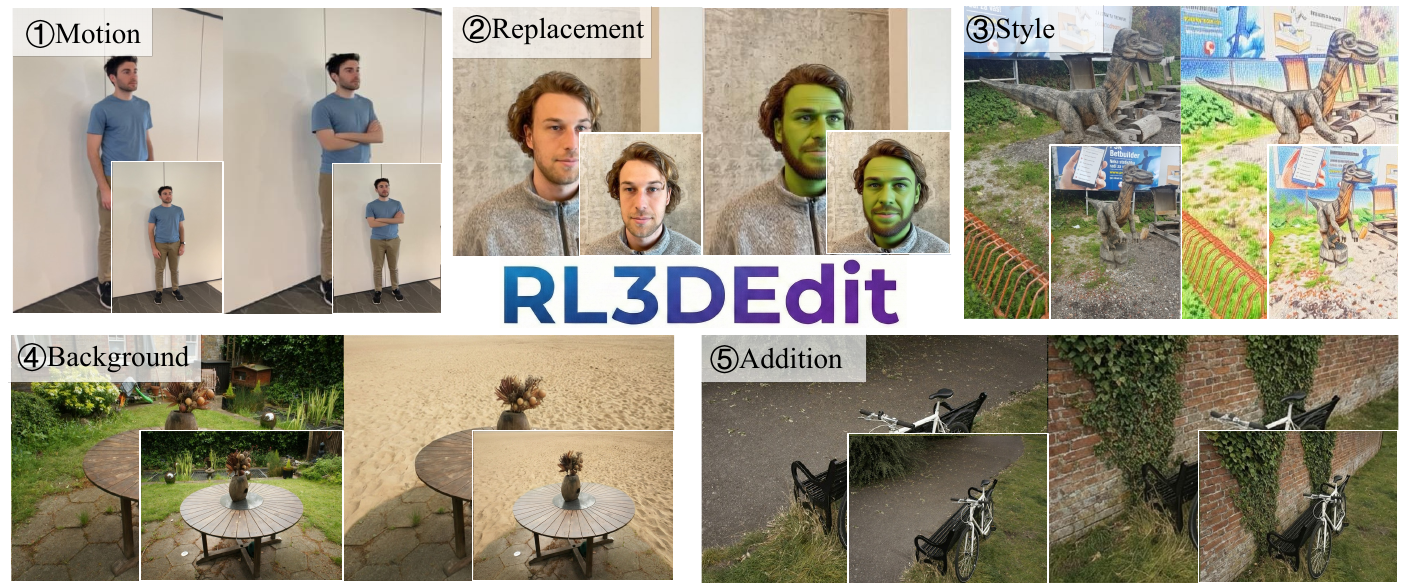}
    \caption{ 
    We propose \textbf{RL3DEdit}, a novel RL-based model for single-pass 3D editing. Our method achieves high-quality results across diverse editing scenarios, including \ding{172}motion edits (cross arms), \ding{173}subject replacement (turn into Hulk), \ding{174}style transfer (colored pencil style), \ding{175}background changes (change to sandy beach), and \ding{176}challenging scene addition (add a brick wall with ivy).
}
\vspace{-3em}
\label{fig:teaser}
\end{figure}
\customfootnotetext{Work done during the internship at AMAP Alibaba Group, and NTU. \textsuperscript{\Letter} Corresponding author; \textsuperscript{{\ding{81}}}Project Leader.}
\begin{abstract}
Leveraging the priors of 2D diffusion models for 3D editing has emerged as a promising paradigm. However, multi-view consistency remains challenging in edited results, and the extreme scarcity of paired 3D-consistent editing data makes supervised fine-tuning (SFT) impractical, despite its effectiveness for editing tasks. In this paper, we observe that, while generating multi-view consistent 3D content is highly challenging, verifying 3D consistency is tractable, naturally positioning reinforcement learning (RL) as a feasible solution. Motivated by this, we propose \textbf{RL3DEdit}, a single-pass framework driven by RL optimization with novel rewards derived from the 3D foundation model, VGGT.
Specifically, we leverage VGGT's robust priors learned from massive real-world data, feed the edited images into it, and utilize the output confidence maps and pose estimation errors as reward signals, effectively anchoring the 2D editing priors onto a 3D-consistent manifold via RL. Extensive experiments demonstrate that RL3DEdit achieves stable multi-view consistency and outperforms state-of-the-art methods in editing quality with high efficiency. To promote the development of 3D editing, we will release the code and model.
\end{abstract}

\section{Introduction}
\label{sec:intro}
3D scene editing plays a pivotal role in applications such as AR/VR and gaming, demanding both high-fidelity semantic manipulation and strict geometric coherence.
To achieve this, generating multi-view consistent images via 2D editors, followed by fine-tuning
3D representations such as 3D Gaussian Splatting (3DGS), has emerged as a
promising paradigm for 3D scene editing. Especially, the recent 2D editing models
(\textit{e.g.}, FLUX-Kontext~\cite{flux_kontext}) achieved remarkable progress in
editing fidelity and multi-image correlated editing~\cite{qwen_image}. Despite this
promising foundation, current 3D editing methods not only underexploit these
powerful models, but also remain limited: (i)
Geometric-conditioned methods~\cite{tinker,gaussctrl} guide 3D consistency with depth
maps of the source image, failing to handle edits involving geometric changes;
(ii) Optimization-based approaches~\cite{in2n, gaussedit} iteratively refine 3D representations with
single-view edits, suffering from low efficiency and blurry artifacts due to
3D-inconsistent signals. (iii) Attention-based models~\cite{dge, vcedit} reproject attention
features across viewpoints, yet struggle to guarantee fine-grained geometric
consistency.

Particularly, attention manipulation is similarly a suboptimal solution in the early stage of 2D editing \cite{prompt2prompt}, but is ultimately surpassed by data-driven supervision \cite{flux_kontext}, suggesting that supervision remains the most effective pathway. However, the extreme scarcity of 3D-consistent editing paired data makes this
approach infeasible. Recently, reinforcement learning (RL)~\cite{grpo} has demonstrated
strong potential in advancing 2D editing models~\cite{flowgrpo}, offering a promising alternative
to address the above challenges. Instead of relying on explicit supervision
from carefully constructed datasets, RL algorithms optimize editing models
using feedback from verifiable reward models (VRMs). In the 3D area, verifying
multi-view consistency is significantly more tractable than generating
consistent images, making RL a promising approach to acquire 3D consistency
without massive paired data.

Motivated by this, the remaining key problem is to identify a robust 3D verifier
that can effectively measure the consistency of editing results. In this paper,
we leverage the 3D foundation model, VGGT~\cite{vggt}, to serve as this verifier. Drawing
an analogy to Score Distillation Sampling (SDS~\cite{dreamfusion})—which leverages a frozen 2D
diffusion model to assess image quality—we argue that a frozen VGGT, trained on
massive real-world 3D data, can provide intrinsic feedback on multi-view
consistency. Through empirical analysis, we observe that its confidence maps,
originally designed for error tolerance, can serve as an effective proxy for evaluating
consistency across views. Furthermore, the camera pose prediction can also
offer explicit feedback on viewpoint arrangement. To 
preserve the 2D editor's editing fidelity, we 
additionally design an anchor strategy that aligns edited outputs with 
high-quality single-view references. Unlike traditional 3D verifiers that are easily ``reward-hacked''~\cite{hack} by textureless or blurry images, the 3D foundation model acts as a robust, geometry-aware reward model by leveraging data-driven priors,
thus providing stable guidance for the RL process.

Based on the above analysis, we propose \textbf{RL3DEdit}, a single-pass framework that augments a 2D editor's 3D consistency prior with RL guided by geometry-aware rewards. To satisfy the prerequisite of RL optimization, we explore base editors' multi-image joint editing capabilities and select FLUX-Kontext, whose inherent cross-view attention lays the foundation for achieving consistency. 
During training, the RL process explores diverse editing candidates and evaluates their consistency via the VGGT reward model. The RL algorithm, GRPO~\cite{grpo}, optimizes the editing model toward predicting 3D consistent results. At inference, the edited multi-view images are reconstructed into 3DGS to yield the final edited 3D scene.

Our method requires no per-scene/prompt fine-tuning and effectively preserves the powerful 2D foundation's capabilities after RL optimization. Trained on limited samples, it successfully learns 3D consistency priors and shows promising generalization to unseen conditions.  Moreover, RL3DEdit can handle geometry-changing instructions, and avoids iterative optimization, enabling single-pass inference that is over \textbf{2$\times$} faster than previous methods.
Extensive experiments confirm that our approach achieves superior editing quality and multi-view consistency with high efficiency. Our contributions can be summarized as follows:
\begin{itemize}
  \item We propose a novel 3D editing RL framework, which effectively empowers 2D editors with 3D capabilities through a tractable 3D consistency verifier, thereby bypassing the scarcity of paired training data. 
  \item We identify that the 3D foundation model with data-driven priors, like VGGT, can serve as a superior verifier. Furthermore, we explore its usage and design tailored rewards to enforce geometric consistency and preserve editing quality.
  \item Our optimization-free RL3DEdit model achieves SoTA 3D editing quality with high efficiency.
\end{itemize}

\section{Related Work}
\subsection{2D Image Editing Models}
In the early stage, 2D editing faced a similar dilemma to current 3D editing: the scarcity of paired editing data. To address this, pioneering works manipulated the cross-attention maps of generative models~\cite{prompt2prompt, attend_and_excite}, achieving fine-grained editing.  Subsequently, with the advent of paired editing datasets, several methods began directly training models to follow explicit editing instructions~\cite{instructpix2pix, cosxl, anyedit}, significantly improving editing quality. Recently, with the support of massive data, high-quality unified editing models have emerged~\cite{flux_kontext, qwen_image}, providing more powerful backbones for 3D editing.
\subsection{3D Editing Models}
3D editing can be broadly categorized into object editing and scene editing. Compared to objects, scenes are significantly more challenging due to the complex backgrounds and multiple entities.

\noindent\textbf{3D Object Editing.}
Recent works~\cite{nano3d, voxhammer, 3deditformer} have achieved significant progress in 3D object editing by constructing dedicated datasets and fine-tuning on the voxel foundation model -- Trellis~\cite{trellis}. However, scene editing is difficult to represent with voxels and poses greater challenges in data construction.

\noindent\textbf{3D Scene Editing.}
Scene editing typically adopts 3DGS or NeRF~\cite{bao2024insertnerf} as the 3D representation. We categorize existing methods into four classes:

\noindent\textit{(i) SDS-based methods}~\cite{dreameditor, progressive3d, instruct3dto3d, ednerf, focaldreamer, gaussedit, dreamcatalyst} leverage diffusion priors via Score Distillation Sampling (SDS) to optimize 3D representations, but typically suffer from blurry textures and over-smoothing.

\noindent\textit{(ii) Iterative optimization-based methods,} pioneered by
IN2N~\cite{in2n} and extended by~\cite{gaussianeditor, instruct3dto3d, local_editing,
view_consistent_3d_editing, gaussctrl, proteusnerf, edit_diffnerf}, alternate between single-view editing and 3D optimization. However, the cross-view information is lacking in the editing process, which leads to inconsistent 3D representation optimization and prolonged editing time.

\noindent\textit{(iii) Gaussian parameter manipulation methods}~\cite{3ditscene, vfeditor} directly modify 3DGS parameters via semantic grounding or parameter-change predictors distilled from 2D editors.  
While efficient, these methods struggle with action-based edits (\textit{e.g.}, ``bowing head''), as such instructions are difficult to translate into explicit Gaussian parameter modifications.

\noindent\textit{(iv) Multi-view consistent editing methods} have become the
dominant paradigm. Similar to 2D editing, early works~\cite{dge, warp, 2405.16823, vicanerf, gaussctrl, vcedit, liao2025thinking} propagate, interact, or project attention features across views, yet accumulate alignment errors in geometrically discontinuous regions. The insufficient constraints also lead to residual fine-grained inconsistencies. Subsequent methods like EditSplat~\cite{editsplat, consistdreamer, 3dego} employ multi-view fusion guidance, conditioning each newly edited view on adjacent/all previously edited ones. But they still produce visible artifacts under instructions that are difficult to quantify precisely
(\textit{e.g.}, ``open mouth''---but how wide?).

Furthermore, most aforementioned methods rely on InstructPix2Pix, strictly limiting their 2D editing upper bound. While Tinker~\cite{tinker} recently adopted the powerful FLUX-Kontext, it relies on depth-map guidance (restricting it to geometry-preserving edits), introduces complex video-model pipelines, and demands massive paired data ($\sim$25K samples). In contrast, our RL3DEdit achieves superior generalization with only 5\% of the data via RL optimization.
\subsection{Reinforcement Learning for 3D Tasks}
To our knowledge, RL3DEdit is the first work to introduce RL into 3D editing. The RL paradigm, particularly GRPO~\cite{grpo}, has achieved remarkable success in LLMs~\cite{deepseekr1} and 2D editing~\cite{flowgrpo}, yet remains underexplored in the 3D domain and is mainly focused on 3D generation. Early works~\cite{mvreward, dreamreward} train human-preference reward models and employ RLHF to better align generated
3D assets with human intent. More recently,
Nabla-R2D3~\cite{nabla_r2d3}/AR3D-R1~\cite{ar3d_r1} adopt GRPO to enhance native 3D diffusion/autoregressive generation pipelines, improving texture detail and overall fidelity. Additionally, several
works~\cite{core3d, scenerevis, metaspatial, respace} leverage RL for 3D understanding and spatial layout, applying ensemble critics and physical feedback to enforce spatial and geometric constraints. 

\begin{figure}[!t]
  \centering
  \hspace*{-1.5em}
  \includegraphics[width=1.05\linewidth]{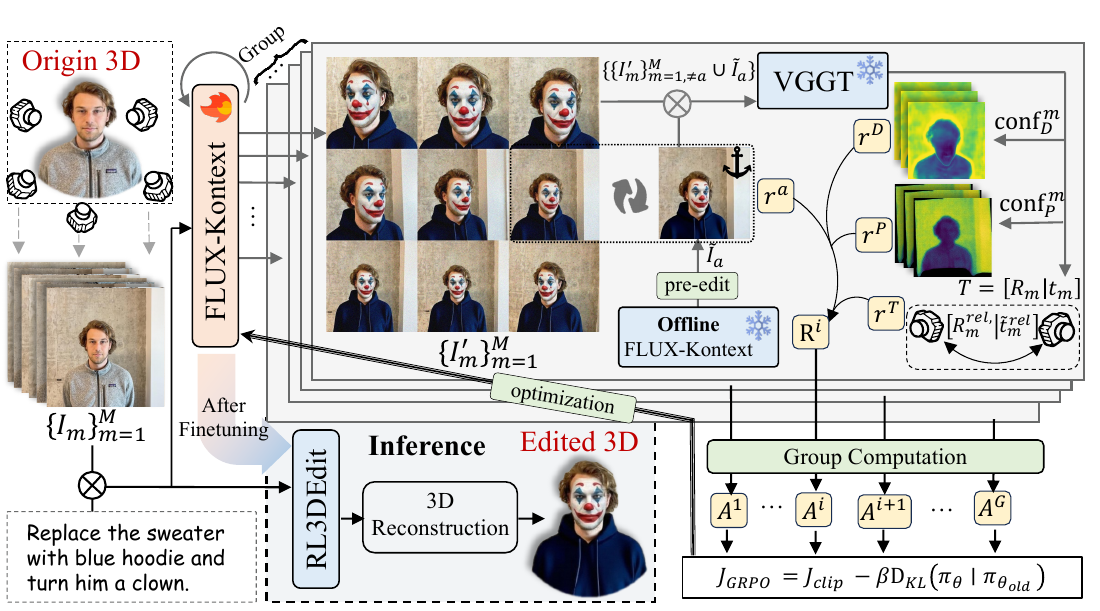}
  \caption{Pipeline of RL3DEdit. Section~\ref{sec:grpo_pipeline} details the pipeline.}
  \vspace{-1.5em}
  \label{fig:pipeline}
\end{figure}
\section{Methods}
\vspace{-0.5em}
\subsection{3D Editing Pipeline with Reinforcement Learning}
\label{sec:grpo_pipeline}
As shown in Fig.~\ref{fig:pipeline}, we construct a novel 3D editing method based on RL. Specifically, given a 3D asset to be edited, we first render it from $M$ viewpoints to obtain $\{I_m\}_{m=1}^{M}$, and feed them simultaneously into a {2D editor} (denoted as $\pi$) for joint multi-view editing. Ideally, at inference, the fine-tuned editor (RL3DEdit) produces multi-view consistent images in a single forward pass, which are then fed into 3DGS reconstruction to obtain the final edited 3D scene, yielding an efficient editing workflow.

Therefore, the core challenge is to equip the 2D editor with a 3D-consistency prior. We address this via RL optimization.
As shown in the upper part of Fig.~\ref{fig:pipeline} (FLUX-Kontext branch), during training, the GRPO algorithm explores a group of $G$ edited results (through $G$ independent inference passes~\cite{flowgrpo}), each containing $M$ edited views $\{I'_m\}_{m=1}^{M}$. 
To explicitly enforce both editing faithfulness and multi-view coherence, we first select an anchor view $I'_a$ and substitute it with our pre-edited anchor image $\tilde I_a$ (detailed in Sec.~\ref{sec:anchor}). Together with the other views, they are fed into a dedicated {3D-aware reward model} (implemented via VGGT~\cite{vggt}), which is designed to assess multi-view consistency with $r^D, r^P, r^T$ and editing quality with $r^a$. These complementary rewards are combined to form the final composite reward ${R^i}$, which effectively guides our optimization toward consistent and high-quality 3D-aware editing. The rewards $\{{R^i}\}_{i=1}^G$ are used to compute the relative advantage via: ${A}^i = (R^i - \mathrm{mean}(\{R^j\}_{j=1}^G))/{\mathrm{std}(\{R^j\}_{j=1}^G)}$, and the model is optimized by maximizing the following objective, which encourages the model to produce high-reward outputs:
$$J(\theta) = J_{\text{clip}}(\theta) - \beta\, D_{\text{KL}}(\pi_\theta \| \pi_{\text{ref}}),$$
where $\pi_\theta/\pi_{\text{ref}}$ denote the fine-tuned/original 2D editors, $J_{\text{clip}}(\theta)$ is computed from ${A}^i$ (see Appendix for details and the formal definition of GRPO). In this way, the model can learn 3D-consistency priors without curated paired supervision.

\begin{figure}[!t]
\centering
\includegraphics[width=\linewidth]{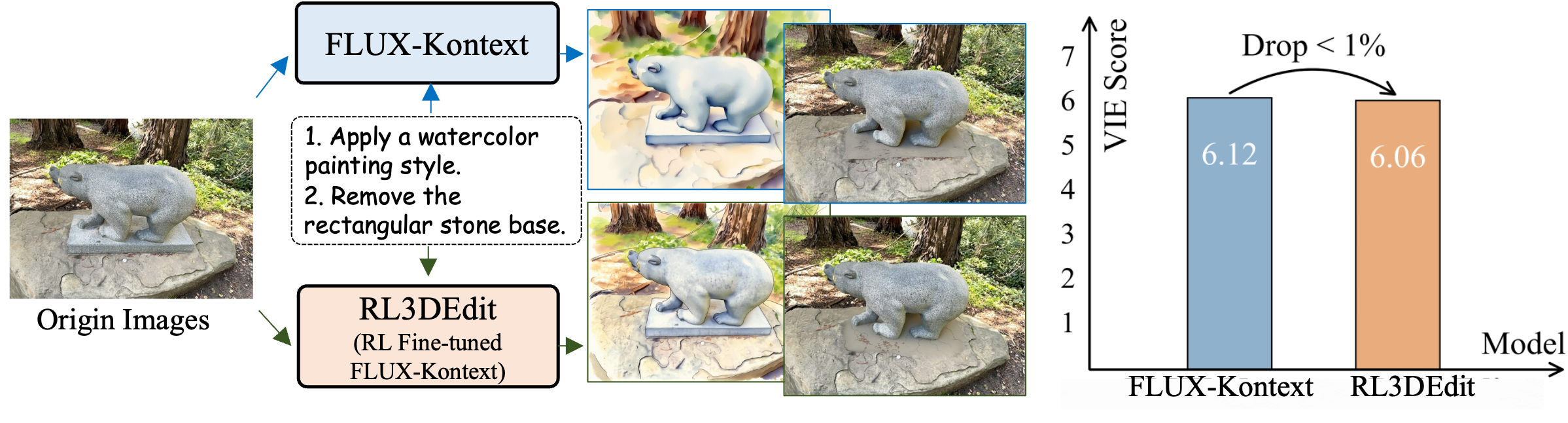}
\caption{Comparison of 2D editing capabilities before and after RL fine-tuning. Left: Visual editing results. Right: Quantitative evaluation using VIEScore~\cite{viescore} on GEdit-Bench-EN~\cite{geditbench_en} ($\uparrow$, detailed in Sec.~\ref{sec:metric}). Both demonstrate that RL3DEdit successfully preserves the original 2D editing fidelity of FLUX-Kontext.}
  \label{fig:vie_score}
  \vspace{-1.5em}
\end{figure}

Compared with previous methods, our method offers \textbf{two} key advantages. On the one hand, verifying multi-view consistency only requires identifying geometric contradictions like ghosting artifacts. This is significantly more tractable than generating consistent images, which requires complex cross-view interactions and relies on scarce multi-view data. This asymmetry 
perfectly aligns with the RL paradigm, where a consistency verifier can naturally serve as the reward model.
On the other hand, 3D editing, like its 2D counterpart, must handle
arbitrary instructions on diverse scenes. Learning such general editing priors in 3D would require prohibitively large paired datasets that currently do not exist. Fortunately, modern 2D foundation editors naturally possess this capability. As shown in Fig.~\ref{fig:vie_score}, the fine-tuned editor
preserves its original 2D editing capability, suggesting that our method primarily augments the 2D model with 3D-consistency priors rather than reshaping it, further highlighting the effectiveness of our method.

In the following, we detail the three core components of our pipeline: the 2D editor (Sec.~\ref{sec:je}), the 3D-aware reward model (Sec.~\ref{sec:consist}),  and the reward design (Sec.~\ref{sec:reward}).

\subsection{Multi-Image Joint Editing}
\label{sec:je}
Multi-view consistent editing requires effective cross-view interaction during the editing process. 
In the RL paradigm, optimization relies on exploring diverse outputs and reinforcing those with high rewards. If a 2D editor processes each view independently, the probability of producing 3D consistent multi-view images is essentially zero. Consequently, RL would have no successful samples to reinforce, causing the optimization to fail. Therefore, multi-image joint editing capability is a necessary prerequisite for RL to be effective.

\begin{figure}[!t]
\centering
\includegraphics[width=\linewidth]{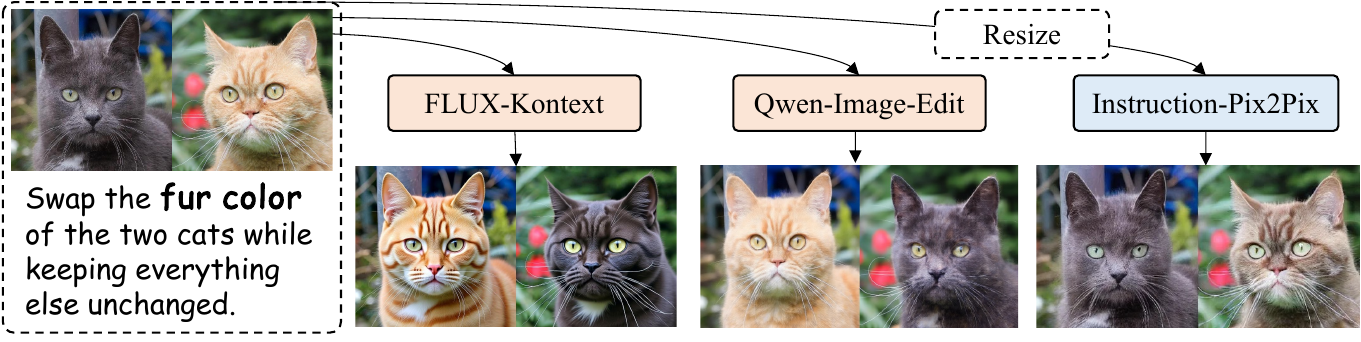}
\caption{Multi-image joint editing comparison. FLUX-Kontext and Qwen-Image-Edit successfully swap the fur colors, while InstructPix2Pix fails due to the lack of cross-view interaction. Moreover, InstructPix2Pix must resize images to low resolution, causing detail loss in multi-image scenarios.}
  \label{fig:ip2p}
  \vspace{-1.5em}
\end{figure}

Previous 3D editing methods typically adopt InstructPix2Pix~\cite{ip2p} as their 2D backbone. However, its low processing resolution and local convolutional operations limit cross-image interaction. As shown in Fig.~\ref{fig:ip2p}, given the instruction to swap the fur colors of two cats, InstructPix2Pix fails to access information from the other images and results in random color changes. In contrast, recent DiT-based models (\textit{e.g.}, FLUX-Kontext~\cite{flux_kontext}, Qwen-Image-Edit~\cite{qwen_edit}) are naturally suited for multi-image editing, as their Transformer architecture enables global attention across all inputs. Specifically, after tokenizing $K$ input images as $\{x^k\}_{k=1}^{K}$, they are concatenated along the sequence dimension as $X=\mathrm{Concat}(x^1,\ldots,x^K)$ and processed with self-attention, enabling attention weights to be distributed across images. This inherent mechanism facilitates efficient cross-view interaction, enabling RL to leverage it to achieve 3D consistency.
Based on this, we adopt FLUX-Kontext as our baseline and further validate the applicability to Qwen-Image-Edit in ablation studies (Sec.~\ref{sec:ablation}).
\subsection{Multi-View Consistent Verification}
\label{sec:consist}
With the RL framework and a promising foundation editor, the remaining question is how to robustly
verify the 3D consistency of edited images. In this paper, we choose a 3D foundation model, VGGT~\cite{vggt}. This is inspired by Score
Distillation Sampling (SDS~\cite{sds}), which employs the 2D foundation model Stable
Diffusion (SD~\cite{sd}) to supervise image quality. SD is trained on high-quality, large-scale images, so feeding it low-quality images produces large loss feedback. Analogously, VGGT is trained on millions of real-world 3D data samples, and
feeding multi-view inconsistent edited images into VGGT can also produce meaningful corresponding feedback.
Below, we explore this feedback form and employ VGGT as our reward model.
Experiments (Sec.~\ref{sec:ablation}) show that such data-prior-based verification is
better than traditional methods (\textit{e.g.}, reprojection warping, Structure-from-Motion)
in avoiding ``reward-hacking''.

\noindent\textbf{3D Foundation Model.   }
\label{sec:prelim_vggt}
Formally, given $N$ views $(I_i)_{i=1}^{N}$ of the same 3D scene, VGGT learns a Transformer $f$ such that $f\big((I_i)_{i=1}^{N}\big)=\big(g_i, D_i, P_i, \tau_i\big)_{i=1}^{N}$, where $g_i, D_i, P_i, \tau_i$ denote the camera parameters, depth map, point map, and tracking features of the $i$-th view, respectively. During training, to handle varying regional difficulty, VGGT jointly predicts confidence maps with error tolerance. The loss function is represented as follows:
$$\mathcal{L}_{\text{geo}} = \|\mathrm{conf}_\text{geo}\odot (\hat {\mathrm{geo}} - \mathrm{geo})\|_1 +
  \|\mathrm{conf}_\text{geo}\odot (\nabla \hat {\mathrm{geo}} - \nabla \mathrm{geo})\|_1 - \alpha\,\|\log
  \mathrm{conf}_\text{geo}\|_1,$$ where $\mathrm{conf}_\text{geo}$ denotes geometric confidence,
including the uncertainty of depth ($\mathrm{conf}_D$) and point ($\mathrm{conf}_P$) predictions.

\noindent\textbf{Confidence Map Analysis.   }
\label{sec:verifier}
To investigate the feedback, as illustrated in Fig.~\ref{fig:analysis}, we render 9 views $\{I_i\}_{i=1}^{9}$ from a 3D asset and independently edit them to obtain $\{I'_i\}$. We then gradually replace $I_i$ with its edited counterpart $I'_i$ to disrupt multi-view consistency, and observe the changes in VGGT's output. Our analysis reveals a strong correlation between the predicted confidence and 3D consistency:
\begin{figure}[!t]
  \centering 
  \includegraphics[width=\linewidth]{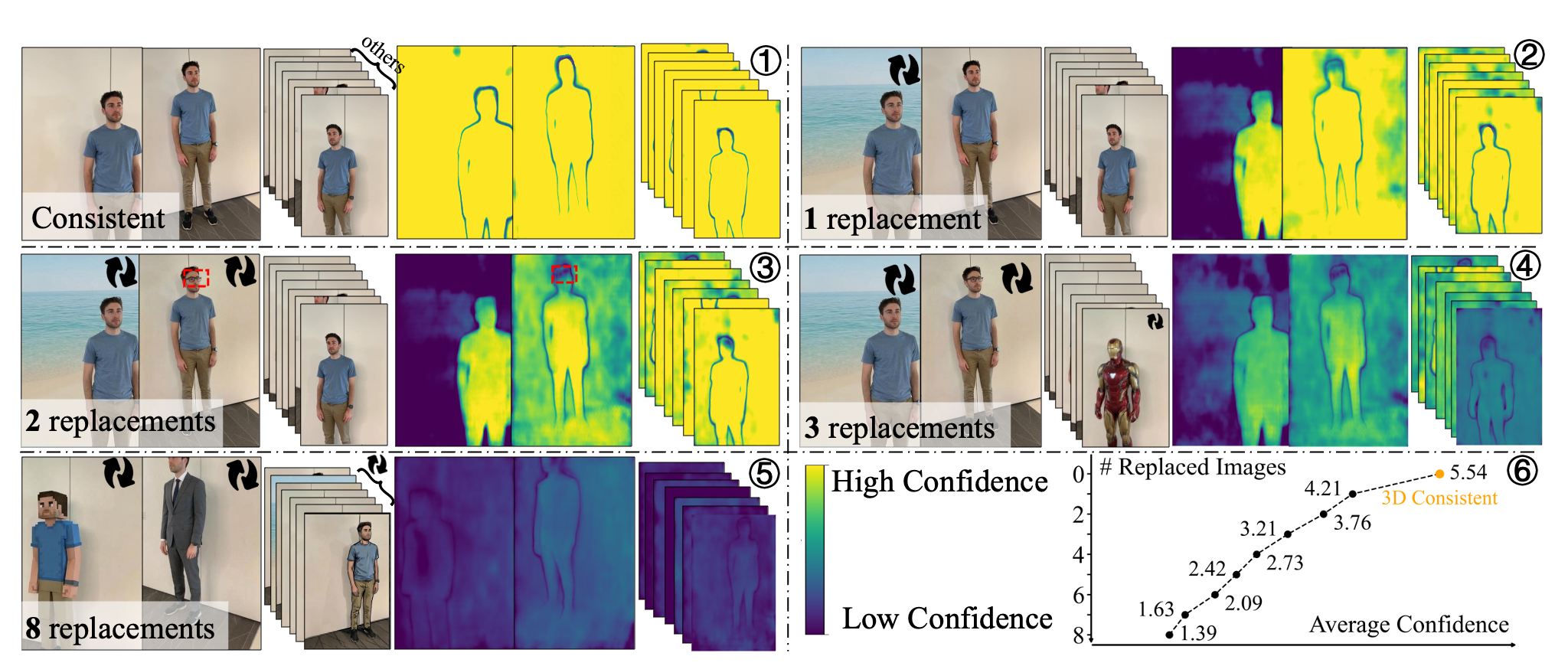}
  \caption{Empirical analysis of VGGT's depth confidence under progressively
    degraded 3D consistency. \ding{172}-\ding{176} visualize VGGT confidence predictions for the \textbf{same} set of 9 views, where individual views are gradually replaced by edited versions. \ding{177} reveals a near-linear correlation between consistency degradation and average confidence. This validates VGGT as the multi-view consistency verifier. Detailed analysis is in Sec.~\ref{sec:verifier}.}
  \label{fig:analysis}
  \vspace{-2.5em}
\end{figure}
\ding{172} When all views are originally 3D-consistent, the
depth confidence is uniformly high. \ding{173} Replacing the background of a single
view, its background confidence drops significantly while the foreground
character remains unaffected. \ding{174} Further adding a pair of glasses in another
view reduces the confidence in the corresponding eye region. \ding{175} Replacing the
person with Iron Man in a third view leads to globally low confidence, as both
the background and foreground now exhibit cross-view inconsistencies. \ding{176}
Finally, when 8 out of 9 views are replaced, although all inputs conceptually
belong to the same scene, they lack mutual 3D consistency, resulting in very
low confidence scores. To quantitatively validate this, we select 100 sets of
9-view data and compute the average confidence under varying replacement
ratios. As shown in \ding{177}, we observe that the predicted confidence decays almost
linearly as consistency decreases. This empirical result confirms that VGGT’s confidence maps are a reliable and interpretable indicator of multi-view consistency. Therefore, VGGT's confidence can
effectively serve as an indicator for 3D consistency and act as a reward.

\subsection{Overview of Reward Model}
\label{sec:reward}
\noindent\textbf{Geometric Rewards.} As discussed above, we use the average depth and point confidence as geometric rewards:
\begin{equation}
  \label{eq:reward}
  r^D = \frac{1}{M}\sum_{m=1}^{M}\mathrm{mean}\big(\mathrm{conf}_D^m\big),
  \qquad
  r^P = \frac{1}{M}\sum_{m=1}^{M}\mathrm{mean}\big(\mathrm{conf}_P^m\big).
\end{equation}

\noindent\textbf{Relative Pose Reward.}
Beyond multi-view consistency, we also consider the arrangement of camera viewpoints.
Considering that editing may alter the absolute camera viewpoint, but the inter-view relationship is still preserved, we use the relative poses between adjacent views to measure camera perspective alignment. Let the VGGT-predicted extrinsics of view $m$ be $T_m=[R_m\,|\,t_m]$, and define the relative transform $T_m^{\mathrm{rel}} = T_{m+1}T_m^{-1}=[R_m^{\mathrm{rel}}\,|\,t_m^{\mathrm{rel}}]$.
We normalize the translation as $\tilde t_m^{\mathrm{rel}}=t_m^{\mathrm{rel}}/\|t_m^{\mathrm{rel}}\|_2$ and use the reference (GT) relative pose $(T_m^{\mathrm{rel}})^{\!*}$, yielding the reward:
\begin{equation}
  r^T = \exp\!\left(-\frac{1}{M-1}\sum_{m=1}^{M-1}
  \Big(\|R_m^{\mathrm{rel}}-(R_m^{\mathrm{rel}})^{\!*}\|_F^2 + \|\tilde t_m^{\mathrm{rel}}-(\tilde t_m^{\mathrm{rel}})^{\!*}\|_2^2\Big)\right).
\end{equation}

\noindent\textbf{Anchor Reward.}
\label{sec:anchor}
As discussed in Sec.~\ref{sec:grpo_pipeline}, RL3DEdit preserves the high editing quality after fine-tuning. This not only benefits from the RL fine-tuning paradigm~\cite{flowgrpo}, but also from the proposed anchor reward. We pre-compute single-image editing results for all views offline using FLUX-Kontext, followed by light quality filtering (empirically, $>$98\% are directly usable). Although these results $\tilde I$ are mutually 3D-inconsistent, they all satisfy FLUX-Kontext's high-quality distribution (\textit{i.e.}, correct semantics, detail preservation, etc.). So we can leverage them independently to guide the RL optimization and preserve FLUX-Kontext's editing fidelity in 3D scenarios.

During training, we randomly sample an anchor index $a\in\{1,\ldots, M\}$ and retrieve $\tilde I_a$ from the offline results (one GRPO-group shares the same index to ensure fair group-reward comparison; randomness is at the sample level). To evaluate the editing quality of $\{I'_m\}_{m=1}^{M}$, we consider two cases: (1) For the anchor view $I'_a$, we directly measure the editing error: 
\begin{equation}
r^a = \exp\big(-\lambda\mathcal{L}_{\text{LPIPS}}(I'_a,\tilde I_a)\big),
\end{equation}
where $\mathcal{L}_{\text{LPIPS}}(\cdot,\cdot)$ denotes the perceptual similarity~\cite{pips}, which aligns well with human perception to prevent edit quality degradation.
(2) For other views, we replace $I'_a$ with $\tilde I_a$ in the multi-view input, feeding $\{I'_1,\ldots,I'_{a-1},\tilde I_a,I'_{a+1},\ldots,I'_M\}$ into VGGT to evaluate consistency. In this way, $I'_{i \ne a}$ are optimized for higher geometric consistency rewards while $I'_a$ is optimized for higher anchor reward, ensuring all views converge toward correct semantics and visual details, thereby preserving FLUX-Kontext's editing fidelity.

\noindent Finally, the $i$-th overall reward in the GRPO-group is defined as:
\begin{equation}
  R^i = w_D r^D + w_P r^P + w_T r^T + w_a r^a,
\end{equation}
which we plug into our framework to optimize FLUX-Kontext.

\section{Experiments}
\vspace{-0.5em}
\subsection{Implementation Details}
We adopt FLUX-Kontext-dev~\cite{flux_kontext} as our baseline and fine-tune it using LoRA with rank 32 and alpha 32. During training, we set $M=9$ and $w_D=w_P=w_T=w_a=0.25$. Following Flow-GRPO~\cite{flowgrpo}, we employ Stochastic Differential Equations (SDE) to enhance exploration randomness, setting the SDE noise level to $0.8$ and the group size to $16$. However, unlike the 6-step denoising exploration used in prior work~\cite{flowgrpo}, our experiments show that 3D consistency demands higher image fidelity, so we adopt a 12-step exploration. To accelerate convergence, we incorporate improvements from MixGRPO~\cite{mixgrpo}, setting the window size to 4.

For training data, we collect 8 scenes from the IN2N~\cite{in2n}, BlendedMVS~\cite{blendedmvs}, and Mip-NeRF360~\cite{mipnerf360} datasets, and construct 7$\sim$9 editing prompts per scene using a VLM~\cite{gemini2_5}, yielding 70 prompts in total (Appendix for details). For each scene, we sample 18$\sim$23 sets of $M$-view images as training samples, where the $M{=}9$ views are evenly sampled around the scene to ensure sufficient visual overlap for consistency evaluation, yielding a total of 1,319 samples. Training was conducted for one epoch on an NVIDIA RTX Pro 6000 GPU and took 42 hours. In inference, we employ 3DGS to reconstruct the edited 3D scene from the multi-view edited images (Appendix for details).

\subsection{Comparison Analysis}
\noindent\textbf{Comparative Models.  } We compare RL3DEdit with the \textit{open-source} SoTA 3D editing methods, including DGE~\cite{dge}, EditSplat~\cite{editsplat}, and GaussCtrl~\cite{gaussctrl}. We note that previous methods adopt InstructPix2Pix~\cite{instructpix2pix} as the 2D editing model; however, as discussed earlier, its limitations in multi-image joint editing make it unsuitable as our baseline. For a fair comparison, we re-implement the strongest baseline, EditSplat, under the same backbone FLUX-Kontext (see Appendix for implementation details) and compare it with our proposed method.  For each scene, we use identical text prompts for editing, then render the edited 3D scene from the same camera poses to obtain results from different methods at consistent novel viewpoints.

\noindent\textbf{Quantitative Comparison and Metrics.   } 
\label{sec:metric}
We evaluate four dimensions: 
(i)~\textit{VIEScore}~\cite{viescore}, a VLM-based metric (GPT-4.1~\cite{gpt}) that jointly evaluates instruction-following and visual quality—we pioneer its usage in 3D editing to provide reproducible, objective assessment free from the subjective variance of user studies; (ii)~\textit{CLIP directional similarity}~\cite{stylegan}, following prior work~\cite{editsplat,gaussedit}; (iii)~\textit{photometric reprojection loss} (Ph-Loss)~\cite{monodepth2} for multi-view consistency (definition in Appendix); and (iv)~\textit{average editing time}.
The test data includes novel views (70 cases), unseen instructions (16 cases), and new scenes (14 cases), totaling 100 test cases (Appendix for detailed split). We note that all prior methods perform per-scene optimization, inherently using the same scene and instruction for both training and testing. Our 70 cases follow the same protocol for fair comparison. For the additional 30 cases, only RL3DEdit operates in a zero-shot setting, while others still optimize on corresponding data. 
As shown in Tab.~\ref{tab:comparison}, RL3DEdit achieves the best performance across all dimensions. It not only achieves a remarkable leap in editing fidelity and semantic alignment (VIEScore of \textbf{5.48} vs. 3.23), but also maintains the lowest Ph-Loss for rigorous 3D consistency. Crucially, this high-quality editing is achieved in just \textbf{1.5 minutes}—over $2\times$ faster than traditional pipelines and over $20\times$ faster than the other FLUX-based baseline.

\begin{table}[!t]
  \centering
  \vspace{-0.5em}
  \caption{Quantitative comparison with SoTA methods. Ph-Loss stands for photometric reprojection loss, $\uparrow/\downarrow$ indicates higher/lower is better. Best results are highlighted in \textbf{bold}. Editing time is tested on an RTX Pro 6000 GPU. Due to space constraints, per-split metrics (in-distribution/zero-shot) and user study are provided in Appendix.}
  \vspace{-0.5em}
  \label{tab:comparison}
  \setlength{\tabcolsep}{4pt}
  \begin{tabular}{@{}lcccc@{}}
    \toprule
    \multirow{2}{*}{Methods} & VIEScore$\uparrow$ & CLIP-dir$\uparrow$ & Ph-Loss$\downarrow$ & Avg. Editing \\
    & \multicolumn{2}{c}{(Novel View Synthesis)} & (Edit-Results) & Time$\downarrow$ \\
    \midrule
    DGE~\cite{dge} & 2.81 & 0.116 & 0.086 & 4min \\
    GaussCtrl~\cite{gaussctrl} & 2.37 & 0.096 & 0.077 & 12min \\
    EditSplat~\cite{editsplat} & 2.72 & 0.121 & 0.081 & 3.5min \\
    EditSplat w/ FLUX-Kontext &3.23&0.125&0.082&40min\\
    \midrule
    RL3DEdit (Ours) & \textbf{5.48} & \textbf{0.147} & \textbf{0.076} & \textbf{1.5min} \\
    \bottomrule
  \end{tabular}
  \vspace{-1em}
\end{table}
\begin{figure}[!t]
  \centering
  \vspace{-0.5em}
  \includegraphics[width=\linewidth]{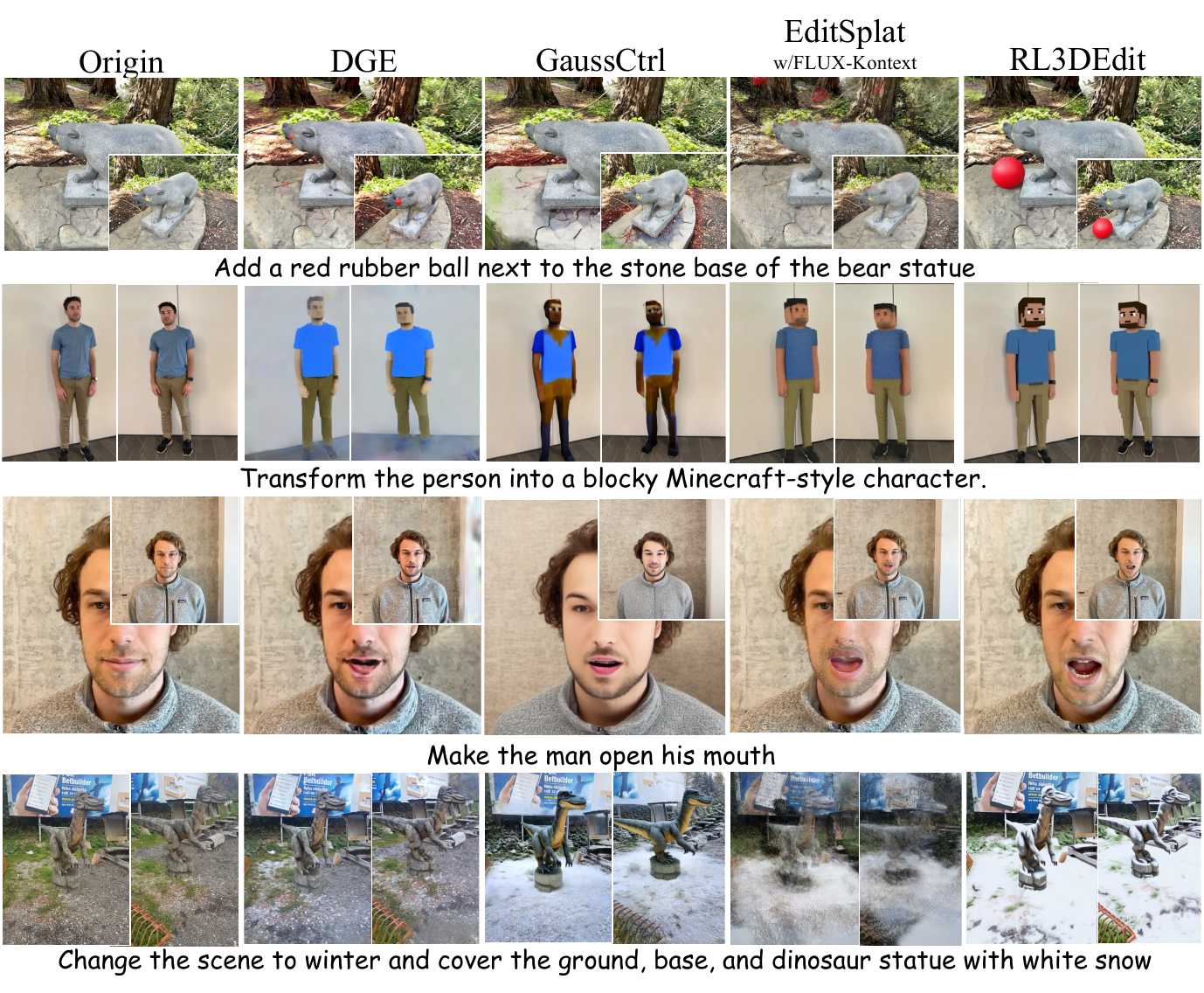}
  \vspace{-2em}
  \caption{Qualitative comparison. RL3DEdit achieves much higher 3D editing quality than other methods, in addition (1st row)/replacement (2nd row)/motion (3rd row)/style (4th row) editing. }
  \label{fig:comparison}
    \vspace{-2em}
\end{figure}

\noindent\textbf{Qualitative Comparison.   }
Metrics alone cannot fully capture editing quality. We therefore present
extensive visual results across
Figs.~\ref{fig:teaser},~\ref{fig:pipeline},~\ref{fig:comparison},~\ref{fig:ablation_reward},
and~\ref{fig:zero_shot}, covering diverse scenes and instructions; additional results are provided in the supplementary
video. Specifically, Fig.~\ref{fig:comparison} presents qualitative comparisons under some challenging instructions. (Row 1) EditSplat/GaussCtrl relies on depth warping/guiding, causing severe failures under geometric-changing prompts. DGE misinterprets the semantics. Only RL3DEdit successfully places the ball in front of the bear. (Row 2) Compared to the blurry results of other methods, RL3DEdit achieves a more realistic Minecraft-style appearance. (Row 3) Motion editing often causes artifacts due to the difficulty in quantifying action magnitude. DGE and EditSplat exhibit artifacts around the mouth, while GaussCtrl alters subject identity. Only RL3DEdit produces correct, high-quality results. (Row 4) For winter scene conversion, DGE merely whitens the scene; GaussCtrl generates snow but alters the dinosaur; EditSplat suffers ghost-artifacts, as massive multi-view inconsistency causes warping failure. Only RL3DEdit achieves semantically accurate editing. These improvements on challenging instructions vividly illustrate the performance advantages shown in Tab.~\ref{tab:comparison}, further demonstrating the effectiveness of our method.

\subsection{Ablation Study}
\label{sec:ablation}
Due to computational constraints, the ablation models are all trained and tested on face scenes with over 200 samples to validate the effectiveness of each component. Experiments confirm that this data scale is sufficient to draw conclusions. Additional ablations on other design details (\textit{e.g.}, anchor selection) are provided in the Appendix.

\begin{figure}[!t]
    \centering
    \includegraphics[width=\textwidth]{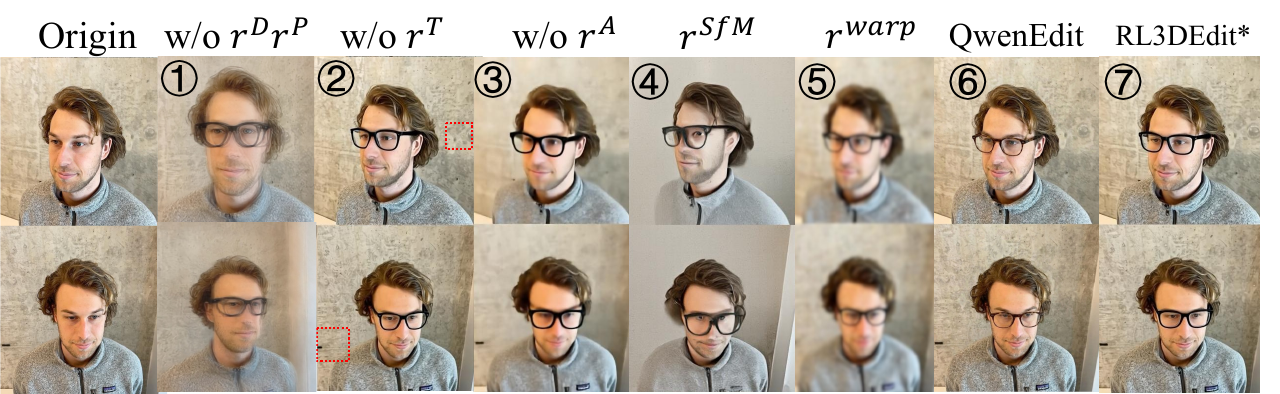}
    \caption{Qualitative ablation results. To better illustrate the causes of degradation, all variants use the same viewpoint for edited results except w/o $(r^D, r^P)$, which uses newly rendered views. In w/o $r^T$, red dashed lines highlight subtle viewpoint shifts compared to the original.}
    \label{fig:ablation_reward}
      \vspace{-1.5em}
\end{figure}

\noindent\textbf{Effect of Rewards.}
Rewards $r^D$ and $r^P$ (depth and point confidence) both constrain 3D consistency, so we ablate them together. \ding{172} Removing $r^D$ and $r^P$ causes significant degradation in both novel-view editing quality and reprojection loss (Tab.~\ref{tab:ablation}). Fig.~\ref{fig:ablation_reward} shows severe ghosting artifacts due to inconsistency, demonstrating the importance of VGGT consistency rewards. \ding{173} Without $r^T$, subtle viewpoint shifts occur, which become evident in reprojection evaluation using GT extrinsics 
(Tab.~\ref{tab:ablation}). Fig.~\ref{fig:ablation_reward} shows displacement in wall details. \ding{174} Without the anchor image to preserve editing priors, outputs degrade toward over-smoothed results (Fig.~\ref{fig:ablation_reward}), primarily because 3D consistency is easier to achieve with low-frequency details, causing RL to optimize in this direction and fail to preserve FLUX-Kontext's editing quality.

\begin{table}[!t]
  \centering
  \caption{Quantitative ablation study on reward components and alternative designs. RL3DEdit* denotes the model trained with the same method but only on face data. The model with $r^{\text{warp}}$ produces blurry images that achieve high consistency but not the desired editing quality. The extension to Qwen-Image-Edit demonstrates that stronger 2D editing models yield better results.}
  \label{tab:ablation}
  \setlength{\tabcolsep}{6pt}
  \begin{tabular}{@{}lcc@{}}
    \toprule
    Methods & VIEScore$\uparrow$ & Ph-Loss$\downarrow$ \\
    \midrule
    \ding{172} w/o $(r^D, r^P)$ & 2.11 & 0.193 \\
    \ding{173} w/o $r^T$ & 4.77 & 0.131 \\
    \ding{174} w/o $r^a$ & 4.34 & 0.091 \\
    \midrule
    \ding{175} replaced w/ $r^{\text{SfM}}$ & 0.97 & 0.201 \\
    \ding{176} replaced w/ $r^{\text{warp}}$ & 1.41 & 0.065 \\
    \midrule
    \ding{177} Qwen-Image-Edit & 5.43 & 0.079 \\
    \ding{178} RL3DEdit* & {5.26} & {0.077} \\
    \bottomrule
  \end{tabular}
    \vspace{-1em}
\end{table}

\noindent\textbf{Effect of Alternative Consistency Verifiers.}
\label{sec:warp}
We compare VGGT against two widely used 3D-consistency measures to 
validate its necessity as a reward.

\noindent\ding{175} \textit{SfM-based reward.} Structure-from-Motion 
requires no learned priors but relies on sparse feature matching 
(definition in Appendix). During RL training, the model quickly learns 
to produce textureless outputs that yield few matchable keypoints, 
trivially satisfying the SfM consistency check while destroying editing 
quality (Fig.~\ref{fig:ablation_reward}, Tab.~\ref{tab:ablation}).

\noindent\ding{176} \textit{Reprojection warping reward.} We compute 
photometric reprojection loss (Ph-Loss)~\cite{monodepth2} using 
VGGT-predicted depth and GT poses as the reward signal (definition in 
Appendix). Tab.~\ref{tab:ablation} shows that the resulting model 
achieves the {lowest} Ph-Loss among all variants, yet produces 
severely blurred outputs (Fig.~\ref{fig:ablation_reward}), confirming 
that Ph-Loss is easily ``reward-hacked'' by low-frequency images. 
It is worth noting that we still adopt Ph-Loss as an evaluation metric 
in Tab.~\ref{tab:comparison}, as it remains effective for measuring 
consistency among sharp images, while any blurriness is penalized by 
VIEScore, ensuring the fairness of our comparison.

Compared to these traditional methods, VGGT is trained on massive real-world 3D images. As a reward function, we think it can optimize consistency while preserving the capability to output high-quality images (further discussed in Appendix). This validates the necessity of using data-driven priors like VGGT as reward models.

\noindent\textbf{Extensions on Qwen-Image-Edit.}
 Both results in Tab.~\ref{tab:ablation} \ding{177} and Fig.~\ref{fig:ablation_reward} \ding{177} demonstrate that our proposed framework can be further enhanced with more powerful editing models. With the rapid advancement of 2D editing models, our framework has strong potential.

\begin{figure}[!t]
    \centering
    \vspace{-1em}
    \includegraphics[width=\textwidth]{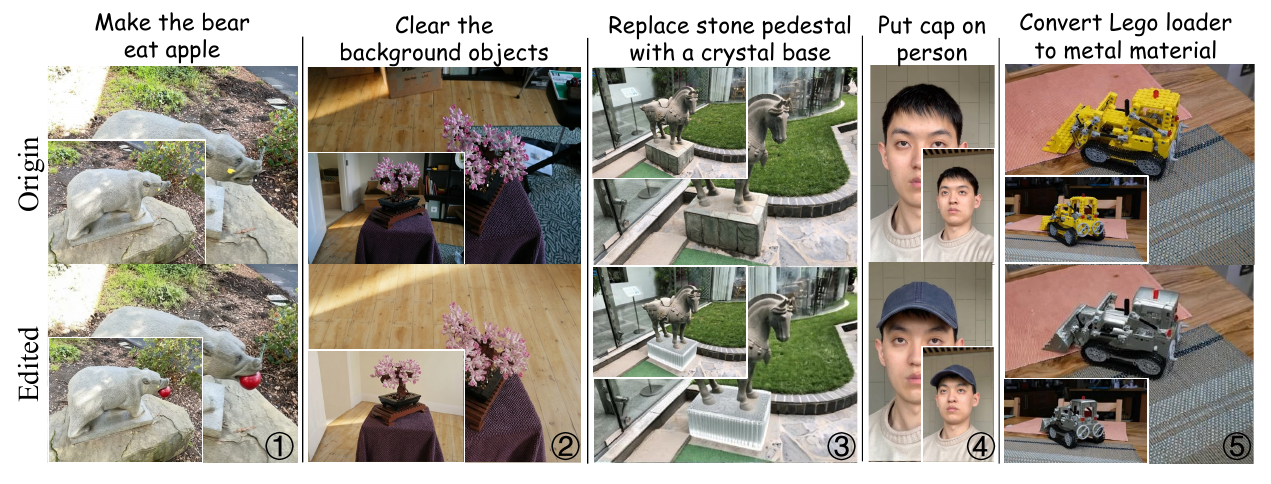}
    \caption{Zero-shot editing results, where \ding{172},\ding{173},\ding{174} are new instructions; \ding{175},\ding{176} are new scenes. More results are available in the supplementary video.}
    \label{fig:zero_shot}
    \vspace{-2em}
\end{figure}

\noindent\textbf{Zero-Shot Generalization.}
\label{sec:zero_shot}
As shown in Fig.~\ref{fig:zero_shot}, by preserving the priors of FLUX-Kontext, our method can generalize to unseen instructions and scenes.

\section{Limitations and Future Work}
\vspace{-0.5em}
\noindent\textbf{Limitations of the 2D Backbone.  }
Since RL3DEdit is fine-tuned on a 2D editor, its performance is bounded by the backbone's inherent constraints. The primary
limitation stems from attention sequence length: multi-view images share the same
token capacity, forcing a trade-off between the number of views and
per-image resolution. However, this is not an inherent defect of our
framework. A natural extension is to use the anchor image as guidance and 
generate edited images in batches to cover more viewpoints, which we leave as future work. Moreover, with the rapid advancement of efficient attention mechanisms---especially streaming and causal attention, which have been successfully applied to long-sequence 3D perception~\cite{stream3r}---context length continues to expand. As demonstrated in Sec.~\ref{sec:ablation}, our method can transfer to other backbones, and this limitation is also expected to diminish as foundation models evolve.

\noindent\textbf{Large Structural Deformations.  }
RL3DEdit may struggle with extremely drastic non-rigid deformations (\textit{e.g.}, ``make the person bow his head''), where the ambiguity in action magnitude leads to subtle multi-view discrepancies that compromise 3D reconstruction. We note that all baselines also fail on such cases, and enhancing generative priors for severe non-rigid deformations remains an open direction.

\noindent\textbf{Training Scale.  } Our training scale is limited, primarily due to the computational overhead of GRPO—each sample requires exploring 16 candidate groups, with each group demanding a 12-step inference pass, resulting in $\sim$2 days of training under our current setup. Nevertheless, RL3DEdit already achieves superior performance, and we believe the method can further improve with increased training scale.
\section{Conclusion}
\vspace{-0.5em}
We present RL3DEdit, an efficient 3D scene editing framework based on RL. Our core insight is that while generating 3D-consistent images is challenging, verifying 3D consistency is tractable, making RL an ideal solution. Building on this, we leverage the 3D foundation model VGGT to construct geometry-aware rewards and employ GRPO to effectively anchor the 2D editor's prior onto the 3D consistency manifold. Experiments demonstrate that RL3DEdit achieves superior editing quality and multi-view consistency with minimal training data, while delivering over 2$\times$ speedup compared to existing methods. Our framework generalizes well and seamlessly transfers to other 2D editing models, offering a new paradigm for future 3D editing research.

\section*{Acknowledgements}
This work was supported by the National Natural Science Foundation of China (NSFC) under Grant (62573039, U2441242).

\bibliographystyle{splncs04}
\bibliography{main}

@String(TOG   = {ACM Trans. Graph.})

@String(TOG   = {ACM TOG})

@inproceedings{in2n,
  title     = {Instruct-nerf2nerf: Editing 3d scenes with instructions},
  author    = {Haque, Ayaan and Tancik, Matthew and Efros, Alexei A and Holynski, Aleksander and Kanazawa, Angjoo},
  booktitle = {Proceedings of the IEEE/CVF international conference on computer vision},
  pages     = {19740--19750},
  year      = {2023}
}

@inproceedings{ip2p,
  title     = {Instructpix2pix: Learning to follow image editing instructions},
  author    = {Brooks, Tim and Holynski, Aleksander and Efros, Alexei A},
  booktitle = {Proceedings of the IEEE/CVF Conference on Computer Vision and Pattern Recognition},
  pages     = {18392--18402},
  year      = {2023}
}

@misc{flux_kontext,
  title         = {FLUX.1 Kontext: Flow Matching for In-Context Image Generation and Editing in Latent Space},
  author        = {Black Forest Labs and Stephen Batifol and Andreas Blattmann and Frederic Boesel and Saksham Consul and Cyril Diagne and Tim Dockhorn and Jack English and Zion English and Patrick Esser and Sumith Kulal and Kyle Lacey and Yam Levi and Cheng Li and Dominik Lorenz and Jonas Müller and Dustin Podell and Robin Rombach and Harry Saini and Axel Sauer and Luke Smith},
  year          = {2025},
  eprint        = {2506.15742},
  archiveprefix = {arXiv},
  primaryclass  = {cs.GR},
  url           = {https://arxiv.org/abs/2506.15742}
}

@misc{tinker,
  title         = {Tinker: Diffusion's Gift to 3D--Multi-View Consistent Editing From Sparse Inputs without Per-Scene Optimization},
  author        = {Canyu Zhao and Xiaoman Li and Tianjian Feng and Zhiyue Zhao and Hao Chen and Chunhua Shen},
  year          = {2025},
  eprint        = {2508.14811},
  archiveprefix = {arXiv},
  primaryclass  = {cs.CV},
  url           = {https://arxiv.org/abs/2508.14811}
}

@article{dge,
  title   = {DGE: Direct Gaussian 3D Editing by Consistent Multi-view Editing},
  author  = {Minghao Chen and Iro Laina and Andrea Vedaldi},
  journal = {arXiv preprint arXiv:2404.18929},
  year    = {2024}
}

@misc{prompt2prompt,
  title         = {Prompt-to-Prompt Image Editing with Cross Attention Control},
  author        = {Amir Hertz and Ron Mokady and Jay Tenenbaum and Kfir Aberman and Yael Pritch and Daniel Cohen-Or},
  year          = {2022},
  eprint        = {2208.01626},
  archiveprefix = {arXiv},
  primaryclass  = {cs.CV},
  url           = {https://arxiv.org/abs/2208.01626}
}

@misc{attend_and_excite,
  title         = {Attend-and-Excite: Attention-Based Semantic Guidance for Text-to-Image Diffusion Models},
  author        = {Hila Chefer and Yuval Alaluf and Yael Vinker and Lior Wolf and Daniel Cohen-Or},
  year          = {2023},
  eprint        = {2301.13826},
  archiveprefix = {arXiv},
  primaryclass  = {cs.CV},
  url           = {https://arxiv.org/abs/2301.13826}
}

@misc{instructpix2pix,
  title         = {InstructPix2Pix: Learning to Follow Image Editing Instructions},
  author        = {Tim Brooks and Aleksander Holynski and Alexei A. Efros},
  year          = {2023},
  eprint        = {2211.09800},
  archiveprefix = {arXiv},
  primaryclass  = {cs.CV},
  url           = {https://arxiv.org/abs/2211.09800}
}

@article{anyedit,
  title   = {AnyEdit: Mastering Unified High-Quality Image Editing for Any Idea},
  author  = {Yu, Qifan and Chow, Wei and Yue, Zhongqi and Pan, Kaihang and Wu, Yang and Wan, Xiaoyang and Li, Juncheng and Tang, Siliang and Zhang, Hanwang and Zhuang, Yueting},
  journal = {arXiv preprint arXiv:2411.15738},
  year    = {2024}
}

@misc{qwen_image,
  title         = {Qwen-Image Technical Report},
  author        = {Chenfei Wu and Jiahao Li and Jingren Zhou and Junyang Lin and Kaiyuan Gao and Kun Yan and Sheng-ming Yin and Shuai Bai and Xiao Xu and Yilei Chen and Yuxiang Chen and Zecheng Tang and Zekai Zhang and Zhengyi Wang and An Yang and Bowen Yu and Chen Cheng and Dayiheng Liu and Deqing Li and Hang Zhang and Hao Meng and Hu Wei and Jingyuan Ni and Kai Chen and Kuan Cao and Liang Peng and Lin Qu and Minggang Wu and Peng Wang and Shuting Yu and Tingkun Wen and Wensen Feng and Xiaoxiao Xu and Yi Wang and Yichang Zhang and Yongqiang Zhu and Yujia Wu and Yuxuan Cai and Zenan Liu},
  year          = {2025},
  eprint        = {2508.02324},
  archiveprefix = {arXiv},
  primaryclass  = {cs.CV},
  url           = {https://arxiv.org/abs/2508.02324}
}

@misc{nano3d,
  title         = {NANO3D: A Training-Free Approach for Efficient 3D Editing Without Masks},
  author        = {Junliang Ye and Shenghao Xie and Ruowen Zhao and Zhengyi Wang and Hongyu Yan and Wenqiang Zu and Lei Ma and Jun Zhu},
  year          = {2025},
  eprint        = {2510.15019},
  archiveprefix = {arXiv},
  primaryclass  = {cs.CV},
  url           = {https://arxiv.org/abs/2510.15019}
}

@misc{voxhammer,
  title         = {VoxHammer: Training-Free Precise and Coherent 3D Editing in Native 3D Space},
  author        = {Lin Li and Zehuan Huang and Haoran Feng and Gengxiong Zhuang and Rui Chen and Chunchao Guo and Lu Sheng},
  year          = {2025},
  eprint        = {2508.19247},
  archiveprefix = {arXiv},
  primaryclass  = {cs.CV},
  url           = {https://arxiv.org/abs/2508.19247}
}

@misc{dreameditor,
  title         = {DreamEditor: Text-Driven 3D Scene Editing with Neural Fields},
  author        = {Jingyu Zhuang and Chen Wang and Lingjie Liu and Liang Lin and Guanbin Li},
  year          = {2023},
  eprint        = {2306.13455},
  archiveprefix = {arXiv},
  primaryclass  = {cs.CV},
  url           = {https://arxiv.org/abs/2306.13455}
}

@misc{3deditformer,
  title         = {Towards Scalable and Consistent 3D Editing},
  author        = {Ruihao Xia and Yang Tang and Pan Zhou},
  year          = {2025},
  eprint        = {2510.02994},
  archiveprefix = {arXiv},
  primaryclass  = {cs.CV},
  url           = {https://arxiv.org/abs/2510.02994}
}

@misc{progressive3d,
  title         = {Progressive3D: Progressively Local Editing for Text-to-3D Content Creation with Complex Semantic Prompts},
  author        = {Xinhua Cheng and Tianyu Yang and Jianan Wang and Yu Li and Lei Zhang and Jian Zhang and Li Yuan},
  year          = {2024},
  eprint        = {2310.11784},
  archiveprefix = {arXiv},
  primaryclass  = {cs.CV},
  url           = {https://arxiv.org/abs/2310.11784}
}

@misc{ednerf,
  title         = {ED-NeRF: Efficient Text-Guided Editing of 3D Scene with Latent Space NeRF},
  author        = {Jangho Park and Gihyun Kwon and Jong Chul Ye},
  year          = {2024},
  eprint        = {2310.02712},
  archiveprefix = {arXiv},
  primaryclass  = {cs.CV},
  url           = {https://arxiv.org/abs/2310.02712}
}

@misc{focaldreamer,
  title         = {FocalDreamer: Text-driven 3D Editing via Focal-fusion Assembly},
  author        = {Yuhan Li and Yishun Dou and Yue Shi and Yu Lei and Xuanhong Chen and Yi Zhang and Peng Zhou and Bingbing Ni},
  year          = {2023},
  eprint        = {2308.10608},
  archiveprefix = {arXiv},
  primaryclass  = {cs.CV},
  url           = {https://arxiv.org/abs/2308.10608}
}

@article{gaussedit,
  title     = {GaussEdit: Adaptive 3D Scene Editing With Text and Image Prompts},
  volume    = {31},
  issn      = {2160-9306},
  url       = {http://dx.doi.org/10.1109/TVCG.2025.3556745},
  doi       = {10.1109/tvcg.2025.3556745},
  number    = {10},
  journal   = {IEEE Transactions on Visualization and Computer Graphics},
  publisher = {Institute of Electrical and Electronics Engineers (IEEE)},
  author    = {Shu, Zhenyu and Yu, Junlong and Chao, Kai and Xin, Shiqing and Liu, Ligang},
  year      = {2025},
  month     = oct,
  pages     = {7769–7780}
}

@misc{dreamcatalyst,
  title         = {DreamCatalyst: Fast and High-Quality 3D Editing via Controlling Editability and Identity Preservation},
  author        = {Jiwook Kim and Seonho Lee and Jaeyo Shin and Jiho Choi and Hyunjung Shim},
  year          = {2025},
  eprint        = {2407.11394},
  archiveprefix = {arXiv},
  primaryclass  = {cs.CV},
  url           = {https://arxiv.org/abs/2407.11394}
}

@misc{dreamfusion,
  title         = {DreamFusion: Text-to-3D using 2D Diffusion},
  author        = {Ben Poole and Ajay Jain and Jonathan T. Barron and Ben Mildenhall},
  year          = {2022},
  eprint        = {2209.14988},
  archiveprefix = {arXiv},
  primaryclass  = {cs.CV},
  url           = {https://arxiv.org/abs/2209.14988}
}

@misc{instruct3dto3d,
  title         = {Instruct 3D-to-3D: Text Instruction Guided 3D-to-3D conversion},
  author        = {Hiromichi Kamata and Yuiko Sakuma and Akio Hayakawa and Masato Ishii and Takuya Narihira},
  year          = {2023},
  eprint        = {2303.15780},
  archiveprefix = {arXiv},
  primaryclass  = {cs.CV},
  url           = {https://arxiv.org/abs/2303.15780}
}

@misc{gaussianeditor,
  title         = {GaussianEditor: Swift and Controllable 3D Editing with Gaussian Splatting},
  author        = {Yiwen Chen and Zilong Chen and Chi Zhang and Feng Wang and Xiaofeng Yang and Yikai Wang and Zhongang Cai and Lei Yang and Huaping Liu and Guosheng Lin},
  year          = {2023},
  eprint        = {2311.14521},
  archiveprefix = {arXiv},
  primaryclass  = {cs.CV},
  url           = {https://arxiv.org/abs/2311.14521}
}

@inproceedings{local_editing,
  title     = {Localized Gaussian Splatting Editing with Contextual Awareness},
  url       = {http://dx.doi.org/10.1109/WACV61041.2025.00509},
  doi       = {10.1109/wacv61041.2025.00509},
  booktitle = {2025 IEEE/CVF Winter Conference on Applications of Computer Vision (WACV)},
  publisher = {IEEE},
  author    = {Xiao, Hanyuan and Chen, Yingshu and Huang, Huajian and Xiong, Haolin and Yang, Jing and Prasad, Pratusha and Zhao, Yajie},
  year      = {2025},
  month     = feb,
  pages     = {5207–5217}
}

@misc{view_consistent_3d_editing,
  title         = {View-Consistent 3D Editing with Gaussian Splatting},
  author        = {Yuxuan Wang and Xuanyu Yi and Zike Wu and Na Zhao and Long Chen and Hanwang Zhang},
  year          = {2025},
  eprint        = {2403.11868},
  archiveprefix = {arXiv},
  primaryclass  = {cs.GR},
  url           = {https://arxiv.org/abs/2403.11868}
}

@misc{gaussctrl,
  title         = {GaussCtrl: Multi-View Consistent Text-Driven 3D Gaussian Splatting Editing},
  author        = {Jing Wu and Jia-Wang Bian and Xinghui Li and Guangrun Wang and Ian Reid and Philip Torr and Victor Adrian Prisacariu},
  year          = {2024},
  eprint        = {2403.08733},
  archiveprefix = {arXiv},
  primaryclass  = {cs.CV},
  url           = {https://arxiv.org/abs/2403.08733}
}

@misc{proteusnerf,
  title         = {ProteusNeRF: Fast Lightweight NeRF Editing using 3D-Aware Image Context},
  author        = {Binglun Wang and Niladri Shekhar Dutt and Niloy J. Mitra},
  year          = {2024},
  eprint        = {2310.09965},
  archiveprefix = {arXiv},
  primaryclass  = {cs.CV},
  url           = {https://arxiv.org/abs/2310.09965}
}

@misc{edit_diffnerf,
  title         = {Edit-DiffNeRF: Editing 3D Neural Radiance Fields using 2D Diffusion Model},
  author        = {Lu Yu and Wei Xiang and Kang Han},
  year          = {2023},
  eprint        = {2306.09551},
  archiveprefix = {arXiv},
  primaryclass  = {cs.CV},
  url           = {https://arxiv.org/abs/2306.09551}
}

@misc{3ditscene,
  title         = {3DitScene: Editing Any Scene via Language-guided Disentangled Gaussian Splatting},
  author        = {Qihang Zhang and Yinghao Xu and Chaoyang Wang and Hsin-Ying Lee and Gordon Wetzstein and Bolei Zhou and Ceyuan Yang},
  year          = {2024},
  eprint        = {2405.18424},
  archiveprefix = {arXiv},
  primaryclass  = {cs.CV},
  url           = {https://arxiv.org/abs/2405.18424}
}

@misc{vfeditor,
  title         = {Variation-aware Flexible 3D Gaussian Editing},
  author        = {Hao Qin and Yukai Sun and Meng Wang and Ming Kong and Mengxu Lu and Qiang Zhu},
  year          = {2026},
  eprint        = {2602.11638},
  archiveprefix = {arXiv},
  primaryclass  = {cs.GR},
  url           = {https://arxiv.org/abs/2602.11638}
}

@misc{2405.16823,
  title         = {Unified Editing of Panorama, 3D Scenes, and Videos Through Disentangled Self-Attention Injection},
  author        = {Gihyun Kwon and Jangho Park and Jong Chul Ye},
  year          = {2024},
  eprint        = {2405.16823},
  archiveprefix = {arXiv},
  primaryclass  = {cs.CV},
  url           = {https://arxiv.org/abs/2405.16823}
}

@misc{vicanerf,
  title         = {ViCA-NeRF: View-Consistency-Aware 3D Editing of Neural Radiance Fields},
  author        = {Jiahua Dong and Yu-Xiong Wang},
  year          = {2024},
  eprint        = {2402.00864},
  archiveprefix = {arXiv},
  primaryclass  = {cs.CV},
  url           = {https://arxiv.org/abs/2402.00864}
}

@misc{vcedit,
  title         = {View-Consistent 3D Editing with Gaussian Splatting},
  author        = {Yuxuan Wang and Xuanyu Yi and Zike Wu and Na Zhao and Long Chen and Hanwang Zhang},
  year          = {2025},
  eprint        = {2403.11868},
  archiveprefix = {arXiv},
  primaryclass  = {cs.GR},
  url           = {https://arxiv.org/abs/2403.11868}
}

@misc{editsplat,
  title         = {EditSplat: Multi-View Fusion and Attention-Guided Optimization for View-Consistent 3D Scene Editing with 3D Gaussian Splatting},
  author        = {Dong In Lee and Hyeongcheol Park and Jiyoung Seo and Eunbyung Park and Hyunje Park and Ha Dam Baek and Sangheon Shin and Sangmin Kim and Sangpil Kim},
  year          = {2025},
  eprint        = {2412.11520},
  archiveprefix = {arXiv},
  primaryclass  = {cs.CV},
  url           = {https://arxiv.org/abs/2412.11520}
}

@misc{consistdreamer,
  title         = {ConsistDreamer: 3D-Consistent 2D Diffusion for High-Fidelity Scene Editing},
  author        = {Jun-Kun Chen and Samuel Rota Bulò and Norman Müller and Lorenzo Porzi and Peter Kontschieder and Yu-Xiong Wang},
  year          = {2024},
  eprint        = {2406.09404},
  archiveprefix = {arXiv},
  primaryclass  = {cs.CV},
  url           = {https://arxiv.org/abs/2406.09404}
}

@misc{3dego,
  title         = {3DEgo: 3D Editing on the Go!},
  author        = {Umar Khalid and Hasan Iqbal and Azib Farooq and Jing Hua and Chen Chen},
  year          = {2024},
  eprint        = {2407.10102},
  archiveprefix = {arXiv},
  primaryclass  = {cs.CV},
  url           = {https://arxiv.org/abs/2407.10102}
}

@misc{flowgrpo,
  title         = {Flow-GRPO: Training Flow Matching Models via Online RL},
  author        = {Jie Liu and Gongye Liu and Jiajun Liang and Yangguang Li and Jiaheng Liu and Xintao Wang and Pengfei Wan and Di Zhang and Wanli Ouyang},
  year          = {2025},
  eprint        = {2505.05470},
  archiveprefix = {arXiv},
  primaryclass  = {cs.CV},
  url           = {https://arxiv.org/abs/2505.05470}
}

@misc{geditbench_en,
  title         = {Step1X-Edit: A Practical Framework for General Image Editing},
  author        = {Shiyu Liu and Yucheng Han and Peng Xing and Fukun Yin and Rui Wang and Wei Cheng and Jiaqi Liao and Yingming Wang and Honghao Fu and Chunrui Han and Guopeng Li and Yuang Peng and Quan Sun and Jingwei Wu and Yan Cai and Zheng Ge and Ranchen Ming and Lei Xia and Xianfang Zeng and Yibo Zhu and Binxing Jiao and Xiangyu Zhang and Gang Yu and Daxin Jiang},
  year          = {2025},
  eprint        = {2504.17761},
  archiveprefix = {arXiv},
  primaryclass  = {cs.CV},
  url           = {https://arxiv.org/abs/2504.17761}
}

@misc{cosxl,
  title         = {Kosmos-G: Generating Images in Context with Multimodal Large Language Models},
  author        = {Xichen Pan and Li Dong and Shaohan Huang and Zhiliang Peng and Wenhu Chen and Furu Wei},
  year          = {2024},
  eprint        = {2310.02992},
  archiveprefix = {arXiv},
  primaryclass  = {cs.CV},
  url           = {https://arxiv.org/abs/2310.02992}
}

@article{hack,
  title   = {Reward Hacking in Reinforcement Learning.},
  author  = {Weng, Lilian},
  journal = {lilianweng.github.io},
  year    = {2024},
  month   = {Nov},
  url     = {https://lilianweng.github.io/posts/2024-11-28-reward-hacking/}
}

@misc{warp,
  title         = {Diffusion-Based Attention Warping for Consistent 3D Scene Editing},
  author        = {Eyal Gomel and Lior Wolf},
  year          = {2024},
  eprint        = {2412.07984},
  archiveprefix = {arXiv},
  primaryclass  = {cs.CV},
  url           = {https://arxiv.org/abs/2412.07984}
}

@article{grpo,
  title     = {DeepSeek-R1 incentivizes reasoning in LLMs through reinforcement learning},
  volume    = {645},
  issn      = {1476-4687},
  url       = {http://dx.doi.org/10.1038/s41586-025-09422-z},
  doi       = {10.1038/s41586-025-09422-z},
  number    = {8081},
  journal   = {Nature},
  publisher = {Springer Science and Business Media LLC},
  author    = {Guo, Daya and Yang, Dejian and Zhang, Haowei and Song, Junxiao and Wang, Peiyi and Zhu, Qihao et al.},
  year      = {2025},
  month     = sep,
  pages     = {633–638}
}

@misc{trellis,
  title         = {Structured 3D Latents for Scalable and Versatile 3D Generation},
  author        = {Jianfeng Xiang and Zelong Lv and Sicheng Xu and Yu Deng and Ruicheng Wang and Bowen Zhang and Dong Chen and Xin Tong and Jiaolong Yang},
  year          = {2025},
  eprint        = {2412.01506},
  archiveprefix = {arXiv},
  primaryclass  = {cs.CV},
  url           = {https://arxiv.org/abs/2412.01506}
}

@inproceedings{vggt,
  title     = {{VGGT}: Visual Geometry Grounded Transformer},
  author    = {Wang, Jianyuan and Chen, Minghao and Karaev, Nikita and Vedaldi, Andrea and Rupprecht, Christian and Novotny, David},
  booktitle = {Proceedings of the IEEE/CVF Conference on Computer Vision and Pattern Recognition},
  year      = {2025}
}

@article{deepseekr1,
  title     = {DeepSeek-R1 incentivizes reasoning in LLMs through reinforcement learning},
  volume    = {645},
  issn      = {1476-4687},
  url       = {http://dx.doi.org/10.1038/s41586-025-09422-z},
  doi       = {10.1038/s41586-025-09422-z},
  number    = {8081},
  journal   = {Nature},
  publisher = {Springer Science and Business Media LLC},
  author    = {Guo, Daya and Yang, Dejian and Zhang, Haowei and Song, Junxiao and Wang, Peiyi and Zhu, et al.},
  year      = {2025},
  month     = sep,
  pages     = {633–638}
}

@misc{mvreward,
  title         = {MVReward: Better Aligning and Evaluating Multi-View Diffusion Models with Human Preferences},
  author        = {Weitao Wang and Haoran Xu and Yuxiao Yang and Zhifang Liu and Jun Meng and Haoqian Wang},
  year          = {2024},
  eprint        = {2412.06614},
  archiveprefix = {arXiv},
  primaryclass  = {cs.CV},
  url           = {https://arxiv.org/abs/2412.06614}
}

@misc{nabla_r2d3,
  title         = {Nabla-R2D3: Effective and Efficient 3D Diffusion Alignment with 2D Rewards},
  author        = {Qingming Liu and Zhen Liu and Dinghuai Zhang and Kui Jia},
  year          = {2025},
  eprint        = {2506.15684},
  archiveprefix = {arXiv},
  primaryclass  = {cs.GR},
  url           = {https://arxiv.org/abs/2506.15684}
}

@misc{dreamreward,
  title         = {DreamReward: Text-to-3D Generation with Human Preference},
  author        = {Junliang Ye and Fangfu Liu and Qixiu Li and Zhengyi Wang and Yikai Wang and Xinzhou Wang and Yueqi Duan and Jun Zhu},
  year          = {2024},
  eprint        = {2403.14613},
  archiveprefix = {arXiv},
  primaryclass  = {cs.CV},
  url           = {https://arxiv.org/abs/2403.14613}
}

@misc{core3d,
  title         = {CoRe3D: Collaborative Reasoning as a Foundation for 3D Intelligence},
  author        = {Tianjiao Yu and Xinzhuo Li and Yifan Shen and Yuanzhe Liu and Ismini Lourentzou},
  year          = {2025},
  eprint        = {2512.12768},
  archiveprefix = {arXiv},
  primaryclass  = {cs.CV},
  url           = {https://arxiv.org/abs/2512.12768}
}

@misc{scenerevis,
  title         = {SceneReVis: A Self-Reflective Vision-Grounded Framework for 3D Indoor Scene Synthesis via Multi-turn RL},
  author        = {Yang Zhao and Shizhao Sun and Meisheng Zhang and Yingdong Shi and Xubo Yang and Jiang Bian},
  year          = {2026},
  eprint        = {2602.09432},
  archiveprefix = {arXiv},
  primaryclass  = {cs.CV},
  url           = {https://arxiv.org/abs/2602.09432}
}

@misc{metaspatial,
  title         = {MetaSpatial: Reinforcing 3D Spatial Reasoning in VLMs for the Metaverse},
  author        = {Zhenyu Pan and Han Liu},
  year          = {2025},
  eprint        = {2503.18470},
  archiveprefix = {arXiv},
  primaryclass  = {cs.CV},
  url           = {https://arxiv.org/abs/2503.18470}
}

@misc{respace,
  title         = {ReSpace: Text-Driven Autoregressive 3D Indoor Scene Synthesis and Editing},
  author        = {Martin JJ. Bucher and Iro Armeni},
  year          = {2025},
  eprint        = {2506.02459},
  archiveprefix = {arXiv},
  primaryclass  = {cs.CV},
  url           = {https://arxiv.org/abs/2506.02459}
}

@misc{ar3d_r1,
  title         = {Are We Ready for RL in Text-to-3D Generation? A Progressive Investigation},
  author        = {Yiwen Tang and Zoey Guo and Kaixin Zhu and Ray Zhang and Qizhi Chen and Dongzhi Jiang and Junli Liu and Bohan Zeng and Haoming Song and Delin Qu and Tianyi Bai and Dan Xu and Wentao Zhang and Bin Zhao},
  year          = {2025},
  eprint        = {2512.10949},
  archiveprefix = {arXiv},
  primaryclass  = {cs.CV},
  url           = {https://arxiv.org/abs/2512.10949}
}

@misc{viescore,
  title         = {VIEScore: Towards Explainable Metrics for Conditional Image Synthesis Evaluation},
  author        = {Max Ku and Dongfu Jiang and Cong Wei and Xiang Yue and Wenhu Chen},
  year          = {2024},
  eprint        = {2312.14867},
  archiveprefix = {arXiv},
  primaryclass  = {cs.CV},
  url           = {https://arxiv.org/abs/2312.14867}
}

@misc{qwen_edit,
  title         = {ReasonEdit: Towards Reasoning-Enhanced Image Editing Models},
  author        = {Fukun Yin and Shiyu Liu and Yucheng Han and Zhibo Wang and Peng Xing and Rui Wang and Wei Cheng and Yingming Wang and Aojie Li and Zixin Yin and Pengtao Chen and Xiangyu Zhang and Daxin Jiang and Xianfang Zeng and Gang Yu},
  year          = {2025},
  eprint        = {2511.22625},
  archiveprefix = {arXiv},
  primaryclass  = {cs.CV},
  url           = {https://arxiv.org/abs/2511.22625}
}

@misc{sds,
  title         = {DreamFusion: Text-to-3D using 2D Diffusion},
  author        = {Ben Poole and Ajay Jain and Jonathan T. Barron and Ben Mildenhall},
  year          = {2022},
  eprint        = {2209.14988},
  archiveprefix = {arXiv},
  primaryclass  = {cs.CV},
  url           = {https://arxiv.org/abs/2209.14988}
}

@misc{sd,
  title         = {High-Resolution Image Synthesis with Latent Diffusion Models},
  author        = {Robin Rombach and Andreas Blattmann and Dominik Lorenz and Patrick Esser and Björn Ommer},
  year          = {2022},
  eprint        = {2112.10752},
  archiveprefix = {arXiv},
  primaryclass  = {cs.CV},
  url           = {https://arxiv.org/abs/2112.10752}
}

@misc{mixgrpo,
  title         = {MixGRPO: Unlocking Flow-based GRPO Efficiency with Mixed ODE-SDE},
  author        = {Junzhe Li and Yutao Cui and Tao Huang and Yinping Ma and Chun Fan and Miles Yang and Zhao Zhong and Liefeng Bo},
  year          = {2025},
  eprint        = {2507.21802},
  archiveprefix = {arXiv},
  primaryclass  = {cs.AI},
  url           = {https://arxiv.org/abs/2507.21802}
}

@misc{blendedmvs,
  title         = {BlendedMVS: A Large-scale Dataset for Generalized Multi-view Stereo Networks},
  author        = {Yao Yao and Zixin Luo and Shiwei Li and Jingyang Zhang and Yufan Ren and Lei Zhou and Tian Fang and Long Quan},
  year          = {2020},
  eprint        = {1911.10127},
  archiveprefix = {arXiv},
  primaryclass  = {cs.CV},
  url           = {https://arxiv.org/abs/1911.10127}
}

@misc{mipnerf360,
  title         = {Mip-NeRF 360: Unbounded Anti-Aliased Neural Radiance Fields},
  author        = {Jonathan T. Barron and Ben Mildenhall and Dor Verbin and Pratul P. Srinivasan and Peter Hedman},
  year          = {2022},
  eprint        = {2111.12077},
  archiveprefix = {arXiv},
  primaryclass  = {cs.CV},
  url           = {https://arxiv.org/abs/2111.12077}
}

@misc{gemini2_5,
  title         = {Gemini 2.5: Pushing the Frontier with Advanced Reasoning, Multimodality, Long Context, and Next Generation Agentic Capabilities},
  author        = {Gheorghe Comanici and Eric Bieber and Mike Schaekermann and Ice Pasupat et al.},
  year          = {2025},
  eprint        = {2507.06261},
  archiveprefix = {arXiv},
  primaryclass  = {cs.CL},
  url           = {https://arxiv.org/abs/2507.06261}
}

@misc{monodepth2,
  title         = {Digging Into Self-Supervised Monocular Depth Estimation},
  author        = {Clément Godard and Oisin Mac Aodha and Michael Firman and Gabriel Brostow},
  year          = {2019},
  eprint        = {1806.01260},
  archiveprefix = {arXiv},
  primaryclass  = {cs.CV},
  url           = {https://arxiv.org/abs/1806.01260}
}

@misc{stream3r,
  title         = {{STream3R}: Scalable Sequential 3D Reconstruction with Causal Transformer},
  author        = {Yushi Lan and Yihang Luo and Fangzhou Hong and Shangchen Zhou and Honghua Chen and Zhaoyang Lyu and Shuai Yang and Bo Dai and Chen Change Loy and Xingang Pan},
  year          = {2025},
  eprint        = {2508.10893},
  archiveprefix = {arXiv},
  primaryclass  = {cs.CV},
  url           = {https://arxiv.org/abs/2508.10893}
}

@misc{pips,
  title         = {The Unreasonable Effectiveness of Deep Features as a Perceptual Metric},
  author        = {Richard Zhang and Phillip Isola and Alexei A. Efros and Eli Shechtman and Oliver Wang},
  year          = {2018},
  eprint        = {1801.03924},
  archiveprefix = {arXiv},
  primaryclass  = {cs.CV},
  url           = {https://arxiv.org/abs/1801.03924}
}

@article{gpt,
  title   = {Gpt-4 technical report},
  author  = {Achiam, Josh and Adler, Steven and Agarwal, Sandhini and Ahmad, Lama and Akkaya, Ilge and Aleman, Florencia Leoni and Almeida, Diogo and Altenschmidt, Janko and Altman, Sam and Anadkat, Shyamal and others},
  journal = {arXiv preprint arXiv:2303.08774},
  year    = {2023}
}

@misc{stylegan,
  title         = {StyleGAN-NADA: CLIP-Guided Domain Adaptation of Image Generators},
  author        = {Rinon Gal and Or Patashnik and Haggai Maron and Gal Chechik and Daniel Cohen-Or},
  year          = {2021},
  eprint        = {2108.00946},
  archiveprefix = {arXiv},
  primaryclass  = {cs.CV},
  url           = {https://arxiv.org/abs/2108.00946}
}

@article{liao2025thinking,
  title   = {Thinking with Camera: A Unified Multimodal Model for Camera-Centric Understanding and Generation},
  author  = {Liao, Kang and Wu, Size and Wu, Zhonghua and Jin, Linyi and Wang, Chao and Wang, Yikai and Wang, Fei and Li, Wei and Loy, Chen Change},
  journal = {arXiv preprint arXiv:2510.08673},
  year    = {2025}
}

@article{anysplat,
  title     = {Anysplat: Feed-forward 3d gaussian splatting from unconstrained views},
  author    = {Jiang, Lihan and Mao, Yucheng and Xu, Linning and Lu, Tao and Ren, Kerui and Jin, Yichen and Xu, Xudong and Yu, Mulin and Pang, Jiangmiao and Zhao, Feng and others},
  journal   = {ACM Transactions on Graphics (TOG)},
  volume    = {44},
  number    = {6},
  pages     = {1--16},
  year      = {2025},
  publisher = {ACM New York, NY, USA}
}

@article{3dgs,
  author  = {Kerbl, Bernhard and Kopanas, Georgios and Leimk{\"u}hler, Thomas and Drettakis, George},
  title   = {3D Gaussian Splatting for Real-Time Radiance Field Rendering},
  journal = {ACM Transactions on Graphics},
  number  = {4},
  volume  = {42},
  month   = {July},
  year    = {2023},
  url     = {https://repo-sam.inria.fr/fungraph/3d-gaussian-splatting/}
}

@misc{zhao2026resilphaseplugandplayphasemapping,
      title={ResilPhase: Plug-and-Play Phase Mapping and Noise-Resilient Macro-Trajectory Extrapolation for Diffusion Acceleration}, 
      author={Qicheng Zhao and Yu Li and Qi Sun and Zheyu Yan},
      year={2026},
      eprint={2606.26769},
      archivePrefix={arXiv},
      primaryClass={cs.AI},
      url={https://arxiv.org/abs/2606.26769}, 
}

@article{sun2025hyperpoint,
  title={HyperPoint: Multimodal 3D Foundation Model in Hyperbolic Space},
  author={Sun, Yiding and Cheng, Haozhe and Lu, Chaoyi and Li, Zhengqiao and Wu, Minghong and Lu, Huimin and Zhu, Jihua},
  journal={Pattern Recognition},
  year={2025},
  publisher={Pergamon}
}

@article{sun2026align,
  title={Align then Adapt: Rethinking Parameter-Efficient Transfer Learning in 4D Perception},
  author={Sun, Yiding and Zhu, Jihua and Cheng, Haozhe and Lu, Chaoyi and Yang, Zhichuan and Chen, Lin and Wang, Yaonan},
  journal={IEEE Transactions on Multimedia},
  year={2026}
}

@article{li2025diffpcn,
  title={DiffPCN: Latent Diffusion Model Based on Multi-view Depth Images for Point Cloud Completion},
  author={Li, Zijun and Yan, Hongyu and Li, Shijie and Luo, Kunming and Lu, Li and Yang, Xulei and Lin, Weisi},
  journal={arXiv preprint arXiv:2509.23723},
  year={2025}
}

@article{gong2025sculpting,
  title={Sculpting features from noise: Reward-guided hierarchical diffusion for task-optimal feature transformation},
  author={Gong, Nanxu and Li, Zijun and Dong, Sixun and Bai, Haoyue and Ying, Wangyang and Wang, Xinyuan and Fu, Yanjie},
  journal={arXiv preprint arXiv:2505.15152},
  year={2025}
}

@inproceedings{bao2026distractorfree,
  title={Distractor-free Generalizable 3D Gaussian Splatting},
  author={Yanqi Bao and Jing Liao and Jing Huo and Yang Gao},
  booktitle={The Fourteenth International Conference on Learning Representations},
  year={2026},
  url={https://openreview.net/forum?id=G33Iemmj3Z}
}

@article{bao2026fisn,
  title={FISN: FInding Spatial Neighborhoods for Generalizable Novel View Synthesis},
  author={Bao, Yanqi and Ding, Tianyu and Huo, Jing and Li, Wenbin and Gao, Yang},
  journal={IEEE Transactions on Visualization and Computer Graphics},
  year={2026},
  publisher={IEEE}
}

@inproceedings{bao2024insertnerf,
  title={Insertnerf: Instilling generalizability into nerf with hypernet modules},
  author={Bao, Yanqi and Ding, Tianyu and Huo, Jing and Li, Wenbin and Li, Yuxin and Gao, Yang},
  booktitle={International Conference on Learning Representations},
  year={2024}
}

@article{jisheng2025decoupled,
  title={Decoupled Seg Tokens Make Stronger Reasoning Video Segmenter and Grounder},
  author={Dang, Jisheng and Wu, Xudong and Yan, Haowen and Zheng, Huicheng and Zheng, Wei-Shi and Lai, Jianhuang and Hu, Bin and Chua, Tat-Seng},
  journal={IEEE Transactions on Pattern Analysis and Machine Intelligence},
  year={2026}
}

\iftrue
\clearpage
\section*{Supplementary Material}
\markboth{Supplementary Material}{Supplementary Material}
\providecommand{\hbAppendixPrefix}{S}
\renewcommand{\thetable}{\hbAppendixPrefix\arabic{table}}
\renewcommand{\thefigure}{\hbAppendixPrefix\arabic{figure}}
\renewcommand{\thesection}{\hbAppendixPrefix\arabic{section}}
\renewcommand{\theequation}{\hbAppendixPrefix\arabic{equation}}
\renewcommand{\theHtable}{supp.\arabic{table}}
\renewcommand{\theHfigure}{supp.\arabic{figure}}
\renewcommand{\theHsection}{supp.\arabic{section}}
\renewcommand{\theHsubsection}{supp.\arabic{section}.\arabic{subsection}}
\renewcommand{\theHsubsubsection}{supp.\arabic{section}.\arabic{subsection}.\arabic{subsubsection}}
\renewcommand{\theHequation}{supp.\arabic{equation}}
\setcounter{table}{0}
\setcounter{figure}{0}
\setcounter{section}{0}
\setcounter{equation}{0}
\setcounter{page}{1}
\providecommand{\theHpage}{}
\renewcommand{\theHpage}{supp.\arabic{page}}

\section{Overview}
\label{sec:supp_overview}
In this supplementary material, we provide additional technical details, experimental results, and discussions that were omitted from the main paper due to space constraints. The detailed table of contents is provided below:

\vspace{0.5em}
\noindent
S1 Overview \dotfill \pageref{sec:supp_overview}\\
S2 Methodology Details \dotfill \pageref{sec:method}\\
\quad S2.1 Formal Definition of GRPO \dotfill \pageref{sec:grpo_formal}\\
\quad S2.2 Definitions of Alternative Rewards and Evaluation Metrics \dotfill \pageref{sec:reward_form}\\
\quad S2.3 3DGS Reconstruction Details \dotfill \pageref{sec:3dgs}\\
\quad S2.4 Baseline Implementation: EditSplat with FLUX-Kontext \dotfill \pageref{sec:baseline}\\
S3 Dataset and Evaluation Setup \dotfill \pageref{sec:data}\\
\quad S3.1 Training Dataset Construction \dotfill \pageref{sec:train_data}\\
\quad S3.2 Detailed Test Data Split \dotfill \pageref{sec:test_split}\\
S4 Additional Experimental Results \dotfill \pageref{sec:exp}\\
\quad S4.1 Per-split Quantitative Metrics \& User Study \dotfill \pageref{sec:per_split}\\
\quad S4.2 Additional Ablation Studies \dotfill \pageref{sec:add_ablation}\\
\quad S4.3 Additional Qualitative Results and Failure Cases \dotfill \pageref{sec:more_qual}\\
S5 Further Discussions \dotfill \pageref{sec:discuss}\\
\quad S5.1 Why VGGT Serves as a Robust Verifier \dotfill \pageref{sec:vggt_discuss}

\section{Methodology Details} \label{sec:method}

\subsection{Formal Definition of GRPO}
\label{sec:grpo_formal}
\subsubsection{Preliminary: FLUX-Kontext Baseline}
\label{sec:prelim_flux}
With the rapid development of deep learning~\cite{jisheng2025decoupled,li2025diffpcn,bao2026fisn,bao2026distractorfree,gong2025sculpting,sun2025hyperpoint,zhao2026resilphaseplugandplayphasemapping,sun2026align}, we adopt FLUX-Kontext as our base 2D editor, which is built upon a Diffusion Transformer (DiT) architecture and the flow-matching paradigm. During training, a trajectory is constructed from the target edited image $x_0$ to pure noise $x_1 \sim \mathcal{N}(0, I)$ over time step $t \in [0,1]$, formulated as $x_t=(1-t)x_0+t x_1$. The network $v_\theta$ is trained to predict the velocity along this path:
\begin{equation}
  \mathcal{L}_\text{FM}(\theta) = \mathbb{E}_{x_0, x_1, t}\big[\|v_\theta(x_t, t, c) - (x_1 - x_0)\|_2^2\big],
\end{equation}
where $c$ denotes the conditions, including editing instructions and reference images. At inference time, sampling is performed by solving the deterministic Ordinary Differential Equation (ODE) starting from Gaussian noise $x_{1}$:
\begin{equation}
  \label{eq:ode}
  \mathrm{d}x_t = v_\theta(x_t, t, c) \mathrm{d}t, \quad t: 1 \rightarrow 0,
\end{equation}
which yields the edited result $x_0$.

Benefiting from the DiT backbone, FLUX-Kontext naturally supports multi-image correlated editing. The tokens of $K$ input images $\{x^k\}_{k=1}^{K}$ are concatenated along the sequence dimension as $X = \mathrm{Concat}(x^1, \dots, x^K)$. Since all image tokens reside in the same sequence, self-attention mechanisms inherently distribute attention weights across different images, thereby enabling effective cross-view interactions.

\subsubsection{Empowering 2D Editors with 3D Capability via GRPO}
\label{sec:prelim_grpo}
To align the 2D editor with 3D consistency using Group Relative Policy Optimization (GRPO)~\cite{grpo}, we formulate the editing model as a conditional policy $\pi_\theta$. Unlike Proximal Policy Optimization (PPO), GRPO introduces a group-relative advantage to stabilize policy updates without requiring a separate Critic (Value Model).

Given a condition $x$ (comprising $M$ source views and an editing instruction), the policy independently generates a group of $G$ candidate outputs $\{y^i\}_{i=1}^{G}$, where each $y^i = \{I'^i_m\}_{m=1}^{M}$ contains $M$ edited views. To meet GRPO's requirement for stochastic exploration, we follow Flow-GRPO~\cite{flowgrpo} by converting the deterministic Flow-ODE (Eq.~\ref{eq:ode}) into an equivalent Stochastic Differential Equation (SDE):
\begin{equation}
  \mathrm{d} x_t = \left(v_\theta(x_t, t, c) + \frac{\sigma_t^2}{2t}\big(x_t + (1-t)v_\theta(x_t, t, c)\big)\right) \mathrm{d}t + \sigma_t \mathrm{d} w_t.
\end{equation}
Specifically, during sampling, the model starts from pure noise and executes 12 denoising steps. After each step, we inject Gaussian noise with a strength of $\sigma_t = 0.8$ to successfully transition from a deterministic ODE to an SDE, augmenting exploration diversity.

Each generated candidate $y^i$ is evaluated by the VGGT-based reward model to obtain a composite reward $R^i$. GRPO then computes the relative advantage $A^i$ for each candidate within the group:
\begin{equation}
  A^i = \frac{R^i - \mathrm{mean}(\{R^j\}_{j=1}^G)}{\mathrm{std}(\{R^j\}_{j=1}^G)}.
\end{equation}
The policy is optimized by maximizing the following objective, which encourages the network to assign higher probabilities to sample trajectories that yielded higher rewards:
\begin{equation}
  J(\theta) = J_{\text{clip}}(\theta) - \beta\, D_{\text{KL}}(\pi_\theta \| \pi_{\text{ref}}),
\end{equation}
where $J_{\text{clip}}(\theta)$ is given by:
\begin{equation}
  J_{\text{clip}}(\theta) = \mathbb{E}_{x}\!\left[ \frac{1}{G}\sum_{i=1}^{G} \min\!\Big( \rho^i(\theta)\,A^i,\; \mathrm{clip}\big(\rho^i(\theta),\,1{-}\epsilon,\,1{+}\epsilon\big)\,A^i \Big) \right].
\end{equation}
Here, $\rho^i(\theta) = \frac{\pi_\theta(y^i \mid x)}{\pi_{\text{old}}(y^i \mid x)}$ is the importance-sampling ratio tracking the change between the new and old policy. By maximizing this objective, the behavior is explicitly guided by the sign of the relative advantage $A^i$: for candidates with higher-than-average rewards ($A^i > 0$), the objective pushes their generation probability $\pi_\theta(y^i \mid x)$ up; conversely, for candidates with lower-than-average rewards ($A^i < 0$), the objective penalizes them and pushes their probability down. The $\mathrm{clip}$ function restricts the ratio $\rho^i(\theta)$ within the interval $[1-\epsilon, 1+\epsilon]$, ensuring that the policy does not update too drastically in any single step, which mitigates destructive updates and provides stable optimization. Finally, the Kullback-Leibler (KL) divergence penalty $D_{\text{KL}}(\pi_\theta \| \pi_{\text{ref}})$ strongly constrains the updated policy $\pi_\theta$ to stay close to the reference policy $\pi_{\text{ref}}$ (the original FLUX-Kontext editor). This effectively prevents the policy from deviating too far, mitigating mode collapse and preserving its powerful 2D image editing priors without curated paired supervision.

\subsection{Definitions of Alternative Rewards and Evaluation Metrics}
\label{sec:reward_form}
\noindent\textbf{Photometric Reprojection Loss (Ph-Loss).}
For reprojection warping, we use depth maps inferred by VGGT. Given an image pair $(I_A, I_B)$ with corresponding inferred depths $(D_A, D_B)$ and camera extrinsics $(T_A, T_B)$, we warp $I_A$ to view $B$ using:
\begin{equation}
   I_{A \rightarrow B} = I_A\big\langle \sigma(T_B T_A^{-1} \cdot D_A \cdot \sigma^{-1}(p_A)) \big\rangle,
\end{equation}
where $p_A$ denotes the 2D pixel coordinates, $\sigma$ denotes camera projection, and $\langle\cdot\rangle$ denotes warp sampling. To handle occlusions and missing regions, the valid region mask $\mathbf{M}$ is determined by depth consistency:
\begin{equation}
   \mathbf{M} = \mathbb{1}\left[ |D_{A \rightarrow B} - D_B| < \tau \right],
\end{equation}
where $D_{A \rightarrow B}$ is warped similarly to $I_{A \rightarrow B}$, and $\tau$ is the depth threshold. Following Monodepth2~\cite{monodepth2}, we use photometric loss within the valid mask to measure consistency. In our evaluation metric, we calculate this over $M-1$ adjacent view pairs $(I_A, I_B)$:
\begin{equation}
  \mathcal{L}_{\text{ph}} = \frac{1}{M-1}\sum_{(A,B)}
  \Big\| \big(0.15 \cdot (I_B - I_{A \to B}) + 0.85 \cdot \mathcal{L}_{\text{SSIM}}(I_B, I_{A \to B}) \big) \odot \mathbf{M} \Big\|_1.
\end{equation}

When using it as an alternative reward (Experiment \ding{176} in paper Table 2), we also average adjacent pairs and define the reward as $r^{\text{warp}} = \exp(-\mathcal{L}_{\text{ph}})$. This reward is unsuitable for training for two reasons: first, lacking GT depth for the newly edited contents leads to accumulated errors during warping; second, blurry image outputs trivially minimize the photometric differences, yielding deceitfully high rewards. Such ``reward-hacking'' severely degrades image quality during optimization.

Despite these limitations as a training reward, it is worth noting that we still adopt $\mathcal{L}_{\text{ph}}$ as a rigorous evaluation metric in paper Table 1. This is because, during evaluation, the photometric loss effectively measures geometric consistency among sharp images, while any potential blurriness collapse is heavily penalized by the independent VIEScore, ensuring a fair and comprehensive comparison between models. 

\vspace{0.5em}
\noindent\textbf{SfM-based Reward.}
Structure-from-Motion (SfM) sequentially performs feature extraction, matching, and triangulation to obtain 3D point clouds, which are then reprojected to corresponding views to assess consistency. Using the GT extrinsics on the $M$ edited views, we define the SfM-based reward as:
\begin{equation}
   r^{\text{SfM}} = (N_{\text{reg}} / M) \cdot \exp\!\big(-\bar{e}_{\text{reproj}}\big),
\end{equation}
where $N_{\text{reg}}$ is the number of successfully registered views and $\bar{e}_{\text{reproj}}$ is the mean reprojection error across all successfully matched keypoints. 

However, because SfM fundamentally relies on sparse feature matching, it is highly susceptible to ``reward-hacking''~\cite{hack}. As shown in paper Figure 7, the RL policy quickly learns to generate textureless, flat, or severely distorted scenes. These textureless images yield very few keypoints, trivially minimizing the reprojection error $\bar{e}_{\text{reproj}}$ and producing spuriously high rewards, but entirely destroying the editing quality. 

\vspace{0.5em}
\noindent\textbf{CLIP Directional Similarity.}
To evaluate how well the edited 3D scene aligns with the semantic instruction without losing the identity of the original scene, we report the CLIP Directional Similarity~\cite{stylegan} in our benchmark (Table 1). This metric measures the cosine similarity between the change in the image embeddings and the change in the text embeddings. Specifically, letting $E_I$ and $E_T$ denote the CLIP image and text encoders respectively, we compute:
\begin{equation}
  \Delta I = E_I(I_{\text{edit}}) - E_I(I_{\text{source}}), \quad \Delta T = E_T(t_{\text{edit}}) - E_T(t_{\text{source}}),
\end{equation}
\begin{equation}
  \text{CLIP-dir} = \frac{\Delta I \cdot \Delta T}{\|\Delta I\| \|\Delta T\|},
\end{equation}
where $I_{\text{source}}$ and $I_{\text{edit}}$ are the rendered views before and after editing, $t_{\text{source}}$ is the text description of the original scene, and $t_{\text{edit}}$ is the provided editing instruction. We average this score across all evaluated novel viewpoints.

\subsection{3DGS Reconstruction Details}
\label{sec:3dgs}
As discussed in paper Sec. 3.1, we employ 3D Gaussian Splatting (3DGS) to reconstruct the final edited 3D scene using the $M=9$ generated multi-view images and the original GT camera parameters. Since iteratively reconstructing using only $9$ sparse views is highly prone to severe overfitting and geometric degeneration, we avoid initializing the 3DGS randomly. Instead, we perform a warm-up initialization using AnySplat~\cite{anysplat}, a feed-forward 3DGS model related to recent generalizable feed-forward 3DGS methods~\cite{bao2026distractorfree,bao2026fisn}, which yields a reliable initial 3D Gaussian state in under $1$ second. Starting from this initialization, we perform iterative optimization using the standard 3DGS~\cite{3dgs} codebase. We fine-tune the initialized Gaussians for $3,\!000$ iterations using a traditional photometric loss on the edited views, guided by the default learning rate schedulers (Taking about 40s). This combination of feed-forward initialization and light-weight iterative optimization successfully yields a high-fidelity edited 3DGS model capable of rendering geometrically consistent images from any novel viewpoint.
\subsection{Baseline Implementation: EditSplat with FLUX-Kontext}
\label{sec:baseline}
As discussed in main paper Sec. 4.2, we re-implement the strongest baseline, EditSplat~\cite{editsplat}, using the same powerful backbone, FLUX-Kontext, for a fair comparison. In its original implementation, EditSplat utilizes InstructPix2Pix~\cite{ip2p} as the 2D diffusion prior. We substitute this single-image editor with FLUX-Kontext while strictly preserving EditSplat's core optimization mechanisms.

Specifically, we retain its sequential \textbf{Multi-view Fusion Guidance (MFG)}. We first obtain initial FLUX-Kontext edits on all source images. Then, utilizing depth-guided iterative alpha blending, these multi-view edits are sequentially projected and fused into each target view to form a fused guidance image $h_M$. During the underlying diffusion process, classifier-free guidance is generalized to simultaneously align the generated output with $h_M$, the original source image $h_S$, and the text prompt $h_T$. 

It is worth noting that, due to the high resolution and slower inference speed of FLUX-Kontext compared to InstructPix2Pix, the overall optimization process of this strong baseline increases from roughly $3.5$ minutes to approximately $40$ minutes per scene.

\section{Dataset and Evaluation Setup} \label{sec:data}
\subsection{Training Dataset Construction}
\label{sec:train_data}
\begingroup
  \renewcommand{\arraystretch}{1.0}
  \footnotesize
  \vspace{-1.5em}
  \begin{longtable}{@{}c | p{0.9\linewidth}@{}}
    \caption{Comprehensive list of 70 editing instructions in our training dataset.}
    \label{tab:all_prompts} \\
    \toprule
    \textbf{Scene} & \textbf{Edit Instruction} \\
    \midrule
    \endfirsthead
    \multicolumn{2}{c}%
    {{\bfseries \tablename\ \thetable{} -- continued from previous page}} \\
    \toprule
    \textbf{Scene} & \textbf{Edit Instruction} \\
    \midrule
    \endhead
    \midrule \multicolumn{2}{r}{{Continued on next page}} \\
    \endfoot
    \bottomrule
    \endlastfoot

    \multirow{9}{*}{\rotatebox{90}{\textbf{Bear}}}
    & Add a red rubber ball next to the stone base of the bear statue. \\
    & Convert the bear statue material to bronze with green patina. \\
    & Remove the rectangular stone base beneath the bear statue. \\
    & Change the bear statue material to dark brown carved wood. \\
    & Replace the gray stone bear with a black-and-white stone panda statue. \\
    & Replace background trees and bushes with dense bamboo forest. \\
    & Change scene lighting to a moonlit night. \\
    & Apply watercolor painting style to the scene. \\
    & Modify the bear statue's pose to sitting on its base. \\
    \midrule
    \multirow{4}{*}{\rotatebox{90}{\textbf{Bicycle}}}
    & Place a canvas backpack on the bench seat near the bicycle handlebars. \\
    & Replace the grass and paved path with dry desert sand. \\
    & Change the black metal bench material to distressed wood. \\
    & Install a blue child safety seat on the bicycle's rear rack. \\
    & Change the scene to winter with snow covered. \\
    & Add a brick wall with ivy along the road, blocking the bushes. \\
    & Convert the entire scene to black-and-white pencil sketch style. \\
    & Remove the white bicycle entirely from the scene. \\
    & Shrink the white bicycle to miniature size at its original position. \\
    \midrule
    \multirow{7}{*}{\rotatebox{90}{\textbf{Bonsai}}}
    & Change all pink LEGO flowers on the bonsai tree to white. \\
    & Change the purple knitted blanket to a deep blue velvet blanket. \\
    & Place a small brown hardcover book on the left corner of the blanket. \\
    & Make the LEGO bonsai tree trunk taller and upright. \\
    & Convert the entire scene to Van Gogh-style oil painting. \\
    & Change the brown wooden stand material to white marble. \\
    & Convert the bonsai into a pine tree bonsai with green needles. \\
    \midrule
    \multirow{10}{*}{\rotatebox{90}{\textbf{Dinosaur}}}
    & Place a red Santa hat on the dinosaur statue's head. \\
    & Change the wooden dinosaur statue material to shiny gold. \\
    & Open the dinosaur statue's mouth. \\
    & Transform into a realistic living dinosaur with lifelike skin texture. \\
    & Change the scene to winter and cover it with white snow. \\
    & Cover the surrounding gravel and grass with thick autumn fallen leaves. \\
    & Convert the entire scene to colored pencil sketch art style. \\
    & Remove all distant buildings, billboards from the background. \\
    & Add a colorful woven scarf around the dinosaur statue's neck. \\
    & Remove the circular stone base beneath the dinosaur statue. \\
    \midrule
    \multirow{9}{*}{\rotatebox{90}{\textbf{Garden}}}
    & Convert the entire garden scene to cartoon style. \\
    & Remove the sphere ball on the base under the table. \\
    & Change the table material to marble with gray veining. \\
    & Replace the vase and dried flowers with a glass vase of red roses. \\
    & Replace the visible grass areas with a pond. \\
    & Replace background and garden with extending sandy beach ground. \\
    & Turn to winter and add white snow to the environment.\\
    & Convert the scene to Monet-style oil painting. \\
    & Place a red-and-white checkered tablecloth on the round table. \\
    \midrule
    \multirow{9}{*}{\rotatebox{90}{\textbf{Person}}}
    & Add fashionable black-framed glasses to the man. \\
    & Make the man cross his arms in front of his chest. \\
    & Make the man laugh heartily. \\
    & Replace the office background with a beach scene. \\
    & Transform the person into a blocky Minecraft-style character. \\
    & Place a ceramic mug in the man's right hand. \\
    & Change the man's clothes to a dark gray business suit. \\
    & Convert the entire scene to comic illustration style. \\
    & Transform the man into Iron Man in full battle armor. \\
    \midrule
    \multirow{4}{*}{\rotatebox{90}{\textbf{Face}}}
    & Add fashionable black-framed glasses to the person. \\
    & Remove the stubble from the person's face. \\
    & Change the gray sweater to a deep blue hoodie. \\
    & Add a baseball cap to the person. \\
    & Convert the portrait to oil painting style. \\
    & Make the man open his mouth. \\
    & Apply clown makeup to the person's face. \\
    & Transform the person into the Hulk. \\
    \midrule
    \multirow{8}{*}{\rotatebox{90}{\shortstack{\textbf{Stone} \textbf{Horse}}}}
    & Transform the stone horse into a realistic living horse with fur texture. \\
    & Add a red traditional Chinese knot tassel around the horse's neck. \\
    & Change the horse statue's surface to blue-and-white porcelain style. \\
    & Colorize the stone horse with brown horse and saddle colors. \\
    & Add yellow fallen leaves on the horse's back and ground. \\
    & Convert the scene to cyberpunk style. \\
    & Switch to nighttime with bright red glowing eyes on the horse statue. \\
    & Switch the scene to rainy weather with wet and reflective surfaces. \\
  \end{longtable}
\endgroup
\subsection{Detailed Test Data Split}
\label{sec:test_split}
\begin{table}[!h]
  \centering
  \vspace{-1.5em}
  \caption{The 16 unseen test instructions utilized for evaluating instruction-level zero-shot generalization on the 8 training scenes.}
  \label{tab:unseen_prompts}
  \renewcommand{\arraystretch}{1.1}
  \resizebox{\linewidth}{!}{
  \begin{tabular}{@{}c | p{0.85\linewidth}@{}}
    \toprule
    \textbf{Scene} & \textbf{Unseen Edit Instruction} \\
    \midrule
    \multirow{2}{*}{\textbf{Bear}}         & Make the bear eat apple \\
                                           & Add a thick layer of snow on the bear statue. \\ \midrule
    \multirow{2}{*}{\textbf{Bicycle}}      & Paint the white bicycle bright red. \\
                                           & Remove the black metal bench. \\ \midrule
    \multirow{2}{*}{\textbf{Bonsai}}       & Clear the background objects and replace with clean wall and floor. \\
                                           & Place a small blue ceramic bird next to the bonsai. \\ \midrule
    \multirow{2}{*}{\textbf{Dinosaur}}     & Replace the wooden dinosaur with a metallic robot dinosaur. \\
                                           & Place a small puddle of water in front of the dinosaur. \\ \midrule
    \multirow{2}{*}{\textbf{Garden}}       & Add a white ceramic rabbit statue on the table. \\
                                           & Change the garden lighting to a warm sunset glow. \\ \midrule
    \multirow{2}{*}{\textbf{Person}}       & Dress the man in a red T-shirt with a white star logo on the chest. \\
                                           & Change the person's hair color to blonde. \\ \midrule
    \multirow{2}{*}{\textbf{Face}}         & Make the person close their eyes. \\
                                           & Add a thick brown beard to the person's face. \\ \midrule
    \multirow{2}{*}{\shortstack{\textbf{Stone} \\ \textbf{Horse}}} & Replace stone pedestal with a crystal base. \\
                                           & Add a heavy metal armor to the horse. \\
    \bottomrule
  \end{tabular}}
  \vspace{-1.5em}
\end{table}
\begin{table}[!h]
  \centering
  \caption{The 14 new scene instructions utilized for evaluating fully zero-shot scene generalization capability across 4 newly introduced test assets.}
  \label{tab:new_scene_prompts}
  \renewcommand{\arraystretch}{1.1}
  \resizebox{\linewidth}{!}{
  \begin{tabular}{@{}c | p{0.85\linewidth}@{}}
    \toprule
    \textbf{Test Scene} & \textbf{New Scenes Edit Instruction} \\
    \midrule
    \multirow{3}{*}{\textbf{Room}}         & Change the wooden floor to marble. \\
                                           & Remove the TV from the room. \\
                                           & Convert the scene to line drawing style. \\ \midrule
    \multirow{4}{*}{\textbf{Stump}}        & Change the scene to winter with snow covering the stump. \\
                                           & Place a small camping lantern on the stump. \\
                                           & Turn the wooden stump into solid stone. \\
                                           & Surround the stump with vibrant spring flowers. \\ \midrule
    \multirow{3}{*}{\textbf{Fangzhou}}     & Put cap on this person. \\
                                           & Add a mustache to this person. \\
                                           & Transform this person into blocky Minecraft style. \\ \midrule
    \multirow{4}{*}{\textbf{Kitchen}}      & Convert Lego loader to metal material. \\
                                           & Place a small LEGO figure next to the loader. \\
                                           & Add a bowl of fruit on the table. \\
                                           & Convert the scene into an underwater environment. \\
    \bottomrule
  \end{tabular}}
  \vspace{-1.5em}
\end{table}
As stated in Section 4.2, our evaluation split consists of 100 testing cases categorized into:
\begin{enumerate}
    \item \textbf{70 Novel Views (In-distribution):} These evaluate view synthesis generalization. They use scenes and instructions seen during training but render them from completely novel viewpoints not used in the optimization trajectories.
    \item \textbf{16 Unseen Instructions (Zero-shot instruction):} Two novel, unseen instructions are drafted for each of the 8 training scenes, evaluating the model's generalization beyond the training instruction. We list them in Table~\ref{tab:unseen_prompts}.
    \item \textbf{14 New Scenes (Zero-shot scene):}  Unseen new scenes (\textit{room, stump, fangzhou, kitchen}) are tested with new instructions to verify the zero-shot generalization capabilities. See detailed instructions in Table~\ref{tab:new_scene_prompts}.
\end{enumerate}

\section{Additional Experimental Results} \label{sec:exp}
\subsection{Per-split Quantitative Metrics \& User Study}
\label{sec:per_split}
As mentioned in paper Table 1, we present quantitative evaluations partitioned into in-distribution and zero-shot settings to better demonstrate the generalization capabilities of our method. Notably, baseline methods necessitate iterative per-scene optimization across all test cases (including both seen and unseen instructions/scenes), inherently remaining within the in-distribution paradigm.

3D editing is essentially a subjective task, so we further conduct a user study. To comprehensively assess both image quality and multi-view consistency simultaneously, we present side-by-side rendered video comparisons of the editing results (our method versus baselines) to 25 participants. The participants are asked to select their overall preferred result based on how well it follows the editing instruction and maintains geometric stability throughout the video. We report the preference rate (\%) for each method.

As shown in Table~\ref{tab:per_split_user}, our method not only achieves state-of-the-art performance on in-distribution data, but also maintains significant advantages in the zero-shot setting over other baselines that directly optimize on the testing scenes. Furthermore, RL3DEdit achieves an overwhelming advantage in the user study, securing the highest preference rate in both seen and unseen scenarios, which strongly corroborates its superior visual quality and 3D consistency.

\begin{table}[!t]
  \centering
  \caption{Per-split quantitative metrics and user study preference rates. ``In-dist.'' refers to the 70 novel-view cases, while ``Zero-shot'' encompasses 30 cases with unseen instructions or newly introduced scenes. Detailed in Sec. \ref{sec:per_split}.}
  \label{tab:per_split_user}
  \renewcommand{\arraystretch}{1.1}
  \resizebox{\linewidth}{!}{
  \begin{tabular}{@{}l ccc ccc@{}}
    \toprule
    \multirow{2}{*}{Methods}
    & \multicolumn{3}{c}{In-dist.\ (70 cases)}
    & \multicolumn{3}{c}{Zero-shot (30 cases)} \\
    \cmidrule(lr){2-4} \cmidrule(lr){5-7}
    & VIEScore$\uparrow$ & User Study$\uparrow$ & Ph-Loss$\downarrow$
    & VIEScore$\uparrow$ & User Study$\uparrow$ & Ph-Loss$\downarrow$ \\
    \midrule
    DGE~\cite{dge}                      & 2.76 & 8.6\%  & 0.086 & 2.92 & 10.0\%  & 0.085 \\
    GaussCtrl~\cite{gaussctrl}          & 2.40 & 4.3\%  & 0.077 & 2.30 & 6.7\%   & 0.078 \\
    EditSplat~\cite{editsplat}          & 2.62 & 5.7\%  & 0.082 & 2.94 & 6.7\%   & 0.080 \\
    \begin{tabular}{@{}l@{}}EditSplat w/ \\[-0.5ex] FLUX-Kontext\end{tabular} & 3.22 & 10.0\% & 0.083 & 3.26 & 16.6\%  & 0.079 \\
    \midrule
    RL3DEdit (Ours)                     & \textbf{5.52} & \textbf{71.4\%} & \textbf{0.075} & \textbf{5.40} & \textbf{60.0\%} & \textbf{0.078} \\
    \bottomrule
  \end{tabular}}
\end{table}
\subsection{Additional Ablation Studies}
\label{sec:add_ablation}
As mentioned in paper Sec.~4.3, due to space constraints, we present additional ablation studies on other design details in this appendix. All ablation models are trained and evaluated on the same Face dataset split following the identical protocol as in the main paper.

\noindent\textbf{Effect of Anchor Selection Strategy.}
To explicitly enforce editing fidelity during the RL process, we utilize a pre-calculated, high-quality single-view edit as an anchor to guide multi-view optimization. We compare two anchor selection strategies: 
(a) \textit{Fixed}: Always choosing the first rendered view as the anchor.
(b) \textit{Random}: Randomly selecting one view per sample during training (our default).

As shown in the top section of Table~\ref{tab:additional_ablation}, the \textit{Random} strategy achieves better performance. This is because the \textit{Random} strategy allows all viewpoints to seamlessly learn the editing prior, ensuring uniform editing quality.

\noindent\textbf{Effect of Denoising Steps.} 
In standard 2D RL editing methodologies, a 6-step denoising process using an SDE formulation is typically sufficient to explore high-quality trajectories~\cite{flowgrpo}. However, in our 3D consistent editing framework, we observe that evaluating geometry consistency demands higher image fidelity and minimal artifacts.
As shown in the middle section of Table~\ref{tab:additional_ablation}, utilizing only 6 denoising steps fails to resolve fine-grained details efficiently, obstructing the VGGT reward model from accurately assessing cross-view 3D consistency, leading to poor Ph-Loss. Increasing the denoising steps to 12 significantly improves the novel view render quality and enables effective 3D alignment. Further increasing to 20 steps yields negligible gains while disproportionately increasing inference and training time. Therefore, we safely adopt 12 steps as our default configuration.

\noindent\textbf{Effect of Reward Weights.}
Our composite reward integrates four critical components: depth consistency ($r^D$), point consistency ($r^P$), relative pose alignment ($r^T$), and anchor-based image quality ($r^a$). Our default configuration equally weights them ($w_D=w_P=w_T=w_a=0.25$). 
We systematically ablate this design by skewing the weight distribution:
(a) \textit{Geometry-heavy}: $w_D=w_P=0.4, w_T=w_a=0.1$.
(b) \textit{Quality-heavy}: $w_a=0.7, w_D=w_P=w_T=0.1$.

The bottom section of Table~\ref{tab:additional_ablation} reveals the delicate balance required for 3D editing. A \textit{Geometry-heavy} setting aggressively forces geometric alignment, resulting in a similar Ph-Loss but inevitably compromising the semantic richness and fine details inherent in the 2D prior (VIEScore drops to 4.31). Conversely, a \textit{Quality-heavy} setting aggressively prioritizes 2D visual appeal. However, this lack of geometric constraint produces severe multi-view inconsistencies. Consequently, the novel view reconstruction suffers from critical artifacts, which destroy the overall fidelity of novel-view editing. The equal weighting strategy successfully negotiates these competing objectives, proving to be the optimal configuration.

\noindent\textbf{Stylized Edits.}
We specifically evaluated performance on stylized editing instructions (\textit{e.g.}, artistic style transfer). While the performance is slightly lower than on realistic edits, it remains acceptable, because the VGGT reward guides 3D consistency over all samples and the consistency prior learned from realistic edits generalizes to stylized ones.

\noindent\textbf{Different Inference Views.}
RL3DEdit is primarily a multi-view framework that allows flexible view handling: fewer views (\textit{e.g.}, 4) can be accommodated by duplicating them into 9-view inputs with similar performance, while scaling to more views requires dedicated designs such as a streaming pipeline, which we leave to future work.

\begin{table}[!t]
  \centering
  \caption{Additional ablation studies on anchor selection strategies, denoising steps, and reward weights. ``Ours (Default)'' denotes the RL3DEdit configuration used in the main paper. All experiments are conducted on the face dataset.}
  \label{tab:additional_ablation}
  \renewcommand{\arraystretch}{1.1}
  \setlength{\tabcolsep}{8pt}
  \begin{tabular}{@{}lcc@{}}
    \toprule
    Configuration & VIEScore$\uparrow$ & Ph-Loss$\downarrow$ \\
    \midrule
    \multicolumn{3}{@{}l}{\textbf{Anchor Selection Strategy}} \\
    \midrule
    Fixed Anchor (1st view)          & 4.54   & 0.087 \\
    Ours (Random, 1 view)            & {5.26}   & {0.077} \\
    \midrule
    \multicolumn{3}{@{}l}{\textbf{Denoising Steps}} \\
    \midrule
    6 steps                          & 3.91   & 0.093 \\
    20 steps                         & 5.28   & 0.078 \\
    Ours (12 steps)                  & {5.26}   & {0.077} \\
    \midrule
    \multicolumn{3}{@{}l}{\textbf{Reward Weights}} \\
    \midrule
    Geometry-heavy ($w_D,w_P=0.4$)   & 4.31   & 0.076 \\
    Quality-heavy ($w_a=0.7$)        & 3.97   & 0.091 \\
    Ours (Equal weights $0.25$)      & {5.26}   & {0.077} \\
    \bottomrule
  \end{tabular}
  \vspace{-1.5em}
\end{table}

\subsection{Additional Qualitative Results and Failure Cases}
\label{sec:more_qual}

We present further qualitative editing results of RL3DEdit in Figure~\ref{fig:supp_qual}. For a more comprehensive multi-view evaluation, we strongly encourage reviewing the supplementary video, which dynamically shows the superior 3D consistency and editing quality of our method.

\noindent\textbf{Failure Cases Analysis.} 
As discussed in Sec.~5, RL3DEdit may struggle with extremely drastic non-rigid deformations (\textit{e.g.}, ``make the person bow his head''), where action magnitude ambiguity leads to multi-view discrepancies that compromise 3D reconstruction. All baselines also fail on such cases.

\begin{figure}[!t]
  \centering
  \includegraphics[width=\linewidth]{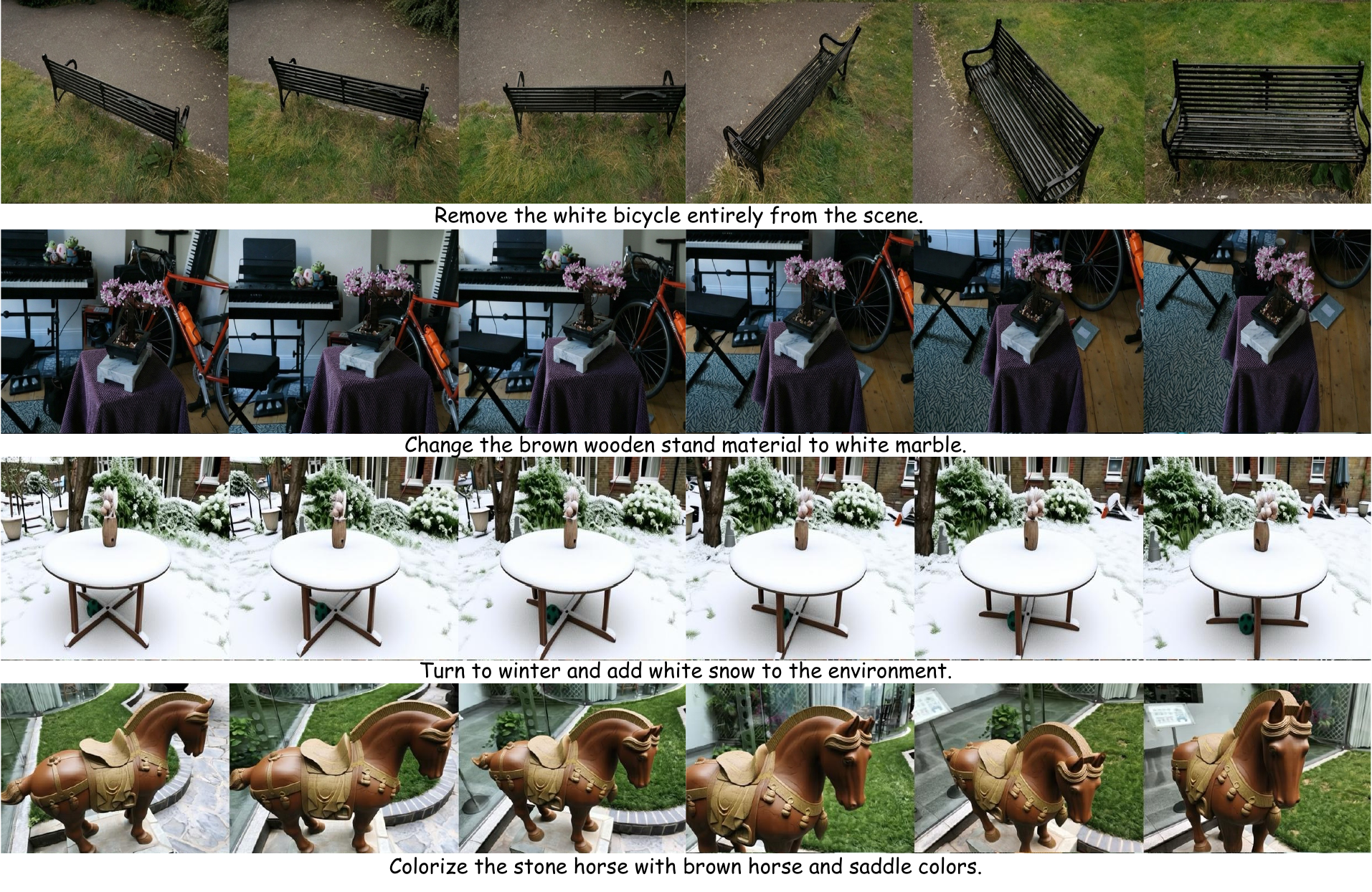} 
  \vspace{-1.5em}
  \caption{Additional qualitative comparisons. (Please zoom in for details and refer to the supplementary video for novel view synthesis).}
  \label{fig:supp_qual}
  \vspace{-1em}
\end{figure}

\section{Further Discussions} \label{sec:discuss}
\subsection{Why VGGT Serves as a Robust Verifier}
\label{sec:vggt_discuss}
As mentioned in paper Sec.~4.3, we argue that VGGT is significantly harder to be ``reward hacked'' than traditional 3D consistency verifiers. For example:
(1) \textbf{SfM} relies on handcrafted feature matching algorithms (e.g., SIFT). If a model collapses and generates textureless images, feature extraction fails, resulting in zero matching points and artificially high SfM rewards. Consequently, an RL policy can maximize this reward by generating distorted or even blank images.
(2) \textbf{Photometric Loss (Ph-Loss)} relies on pixel-level photometric comparisons. If the generated images are excessively blurry or dominated by low-frequency noise, spatially adjacent pixels become nearly identical, leading to artificially minimized errors. Therefore, this metric cannot distinguish between meaningful, high-quality edits and low-frequency artifacts.
\begin{figure}[!t]
  \centering
  \includegraphics[width=0.8\linewidth]{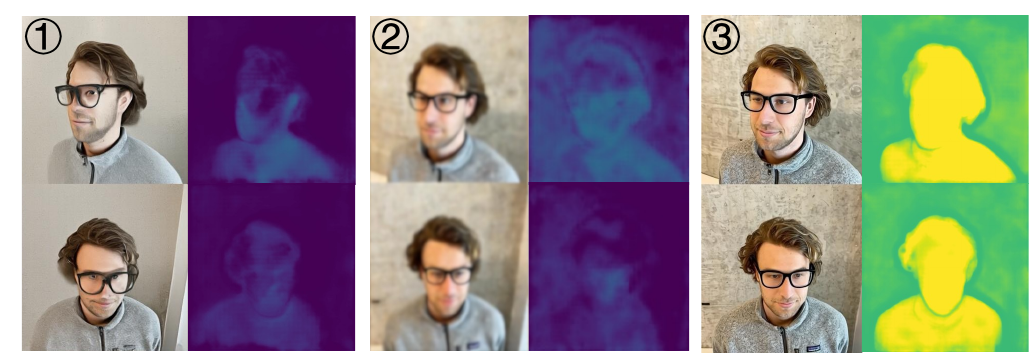} 
  \vspace{-0.5em}
  \caption{Visualization of VGGT confidence maps. For blurry or distorted inputs (\ding{172} and \ding{173}), VGGT assigns lower confidence (darker regions). In \ding{174}, high-fidelity regions yield high confidence (the wall textures naturally receive slightly lower confidence due to the increased consistency difficulty.) This ensures that the reward effectively penalizes low-quality or ``reward-hacked'' outputs.}
  \label{fig:supp_analysis}
  \vspace{-1em}
\end{figure}
Conversely, VGGT possesses the following advantages:
(a) \textbf{Implicit Real-World Priors.} VGGT is trained on millions of multi-view sets of real-world scenes. Through this extensive pre-training, it learns a robust implicit prior regarding the natural structure and appearance of real-world imagery.
(b) \textbf{Fidelity-Conditioned Confidence.} As shown in Fig~\ref{fig:supp_analysis}, when evaluating blurry, textureless, or highly distorted images, VGGT inherently outputs lower confidence maps.
(c) \textbf{Dual Constraint Synergy.} The VGGT reward functions as a powerful ``dual constraint.'' It simultaneously demands strict geometric consistency across viewpoints while explicitly enforcing that the generated images remain natural and realistic.


\fi
\end{document}